%% file: main.tex
\documentclass[10pt,journal,compsoc]{IEEEtran}
\usepackage{times}
\usepackage{epsfig}
\usepackage{graphicx}
\usepackage{amsmath}
\usepackage{amssymb}

\usepackage{multirow}
\usepackage{makecell}
\usepackage{caption}
\usepackage{subcaption}
\usepackage{algorithm}
\usepackage{algpseudocode}
\usepackage{algorithmicx}
\usepackage{float}

\usepackage{color}
\usepackage{xcolor,colortbl}
\usepackage{framed}
\usepackage{enumitem}
\usepackage{booktabs}

\usepackage[flushleft]{threeparttable}

\usepackage{bbm}

\definecolor{shadecolor}{rgb}{0.9,0.9,0.9}
\definecolor{Gray}{gray}{0.9}

\def\eg{\emph{e.g. }} 
\def\ie{\emph{i.e. }}

\def\wrt{w.r.t. }

\algnewcommand\algorithmicinput{\textbf{Input:}}
\algnewcommand\INPUT{\item[\algorithmicinput]}

\usepackage[pagebackref=true,breaklinks=true,letterpaper=true,colorlinks,bookmarks=false]{hyperref}

\ifCLASSOPTIONcompsoc
  \usepackage[nocompress]{cite}
\else
  \usepackage{cite}
\fi
\ifCLASSINFOpdf
\else
\fi
\begin{document}
\title{Open Long-Tailed Recognition \\ In A Dynamic World}

\author{Ziwei~Liu$^{*}$,
        Zhongqi~Miao$^{*}$,
        Xiaohang~Zhan,
        Jiayun~Wang,
        Boqing~Gong,
        and~Stella~X.~Yu.
        % ~\IEEEmembership{Member,~IEEE}% <-this % stops a space
\IEEEcompsocitemizethanks{\IEEEcompsocthanksitem Z. Liu is with Nanyang Technological University. Z. Miao, J. Wang and S. Yu are with the University of California, Berkeley and International Computer Science Institute. X. Zhan is with The Chinese University of Hong Kong. B. Gong is with Google Inc.\protect
% E-mail: see http://www.michaelshell.org/contact.html
\IEEEcompsocthanksitem $^{*}$Equal contribution.}% <-this % stops an unwanted space
% \thanks{Manuscript received April 19, 2005; revised August 26, 2015.}
}


% \markboth{Journal of \LaTeX\ Class Files,~Vol.~14, No.~8, August~2015}%
\markboth{IEEE TRANSACTIONS ON PATTERN ANALYSIS AND MACHINE INTELLIGENCE}%
{Shell \MakeLowercase{\textit{et al.}}: Bare Demo of IEEEtran.cls for Computer Society Journals}



\IEEEtitleabstractindextext{%

\input{sections/abstract.tex}

% \begin{abstract}
% The abstract goes here.
% \end{abstract}

}

\maketitle

\IEEEdisplaynontitleabstractindextext

% \fi
\IEEEpeerreviewmaketitle

\input{sections/intro.tex}

\input{sections/relatedwork.tex}

\input{sections/approach.tex}

\input{sections/experiments.tex}

\input{sections/conclusion.tex}

{
\bibliographystyle{IEEEtran}
\bibliography{reference}
}

\input{sections/bio.tex}

\end{document}

%% file: sections/abstract.tex
%------------------------------------------------------------------------
\begin{abstract}

Real world data often exhibits a long-tailed and open-ended (\ie with unseen classes) distribution. A practical recognition system must balance between majority (head) and minority (tail) classes, generalize across the distribution, and acknowledge novelty upon the instances of unseen classes (open classes).  We define Open Long-Tailed Recognition++ (OLTR++) as learning from such naturally distributed data and optimizing for the classification accuracy over a balanced test set which includes both known and open classes.
OLTR++ handles imbalanced classification, few-shot learning, open-set recognition, and active learning in one integrated algorithm, whereas existing classification approaches often focus only on one or two aspects and deliver poorly over the entire spectrum.  
The key challenges are: 1) how to share visual knowledge between head and tail classes, 2) how to reduce confusion between tail and open classes, and 3) how to actively explore open classes with learned knowledge.
Our algorithm, OLTR++, maps images to a feature space such that visual concepts can relate to each other through a memory association mechanism and a learned metric (dynamic meta-embedding) that both respects the closed world classification of seen classes and acknowledges the novelty of open classes. 
% Our so-called dynamic meta-embedding combines direct image features and associated memory features, with the embedding norms indicate the familiarity to known classes \textcolor{red}{to do what?}.
%
Additionally, we propose an active learning scheme based on visual memory, which learns to recognize open classes in a data-efficient manner for future expansions.
On three large-scale open long-tailed datasets we curated from ImageNet (object-centric), Places (scene-centric), and MS1M (face-centric) data, as well as three standard benchmarks (CIFAR-10-LT, CIFAR-100-LT, and iNaturalist-18), our approach, as a unified framework, consistently demonstrates competitive performance. Notably, our approach also shows strong potential for the active exploration of open classes and the fairness analysis of minority groups. 
    
\end{abstract}


%% file: sections/intro.tex
% long-tail & open-world is the real-world scenario
% curated dataset as a contribution
% Natural application is open long-tail; we don't need a motivation from human intelligence
% not strike the balance, but achieve across-the-board improvements
% "compensate" -> another word
% remove the quotes on "direct feature"
% remove technical details such as SGD
% draw relations to previous works on attention and memory

% define the problem: open long-tail recognition
% why we do this: (1) it's natural; (2) existing methods deal with one aspect
% what are the challenges: (1) knowledge transfer; (2) forgetting; (3) few-shot can be obstructed into open classes, we still need to differentiate the tail classes and novel classes.
% how we do this (layout the landscape, to the layman, technical insight)
% what is the significance

%------------------------------------------------------------------------
\IEEEraisesectionheading{\section{Introduction}}

\IEEEPARstart{O}{ur} visual world is inherently long-tailed and open-ended~\cite{reed2001pareto, liu2020open}, with a few common visual categories (i.e., head classes) and many more relatively rare categories (i.e., tail classes). At the same time, new visual concepts constantly emerge as we navigate in an open world (i.e., open classes).

% However, existing works are mostly developed in a closed setting no matter for the head~\cite{deng2009imagenet, lin2014microsoft} or the tail~\cite{vinyals2016matching, lake2015human}. 
% Namely, all test instances are drawn from the seen classes of which there exist examples to train a classifier. 

%However, existing works are mostly developed for only one aspect, no matter for the head~\cite{deng2009imagenet, lin2014microsoft}, the tail~\cite{vinyals2016matching, lake2015human} or in a closed setting~\cite{wang2017learning}. Traditional deep models~\cite{krizhevsky2012imagenet, he2016deep} are adept at assimilating big data of the head classes and the recently introduced few-shot learning methods~\cite{snell2017prototypical, hariharan2017low} demonstrate promising results in the tail regime of low-density data, but they both fail to consider the full spectrum as shown in Figure~\ref{fig:intro}. 

\begin{figure}[t]
  \centering
  \includegraphics[width=0.48\textwidth]{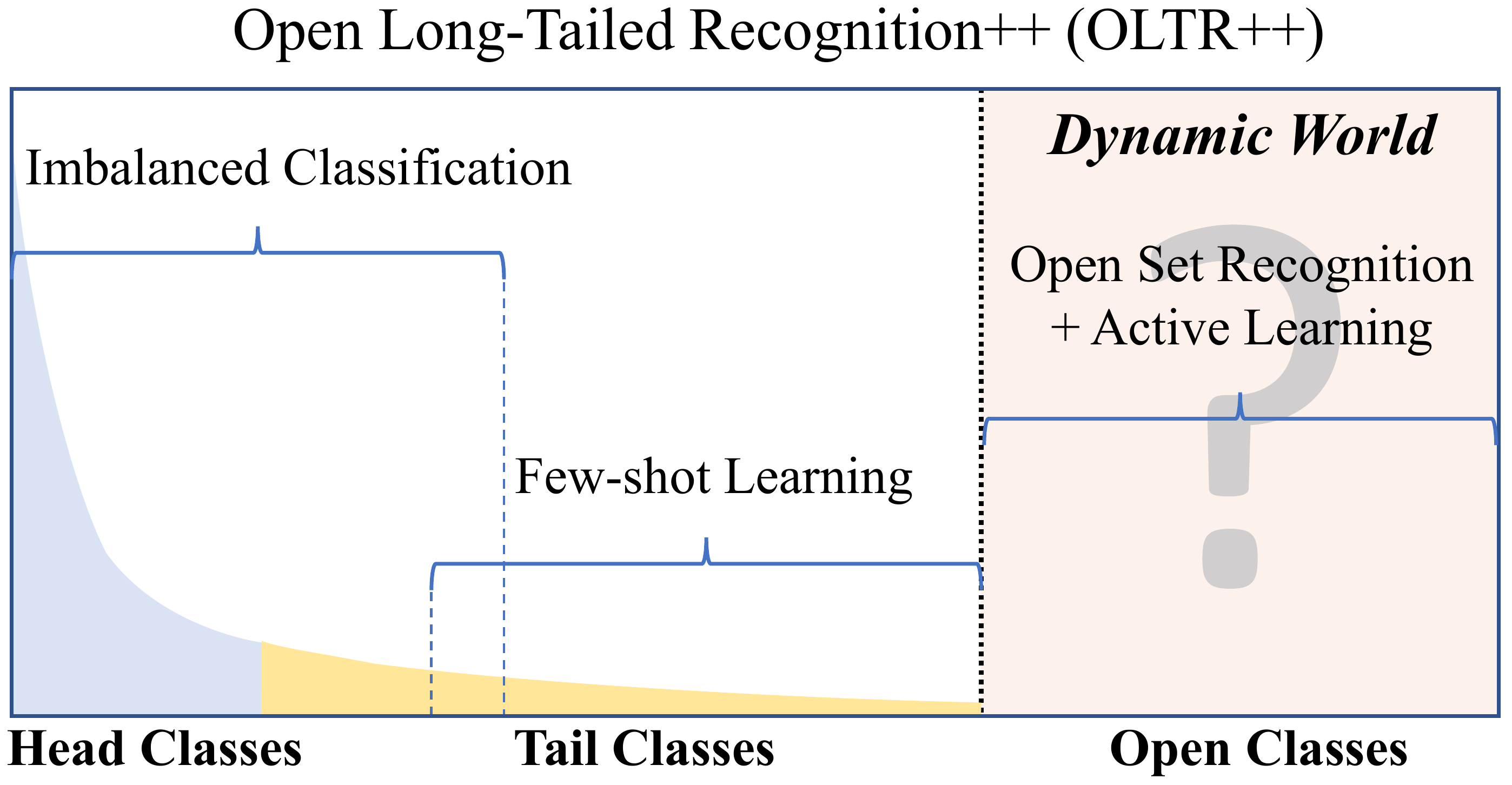}
%  \caption{\footnotesize \textbf{Illustration of long-tailed data with open world deployment.} The visual world is inherently long-tailed and open-set. In this work, we formally design the problem of open long-tailed recognition (OLTR) as optimizing for the overall accuracy of a naturally-distributed dataset with the presence of open classes.}
    \caption{\textbf{OLTR++.} Our task of open long-tailed recognition++ learns from long-tail distributed training data in an open world and deals with imbalanced classification, few-shot learning, open-set recognition, and active learning over the entire spectrum.}
\label{fig:intro}
\end{figure}

\begin{figure*}
  \centering
  \includegraphics[width=1.0\textwidth]{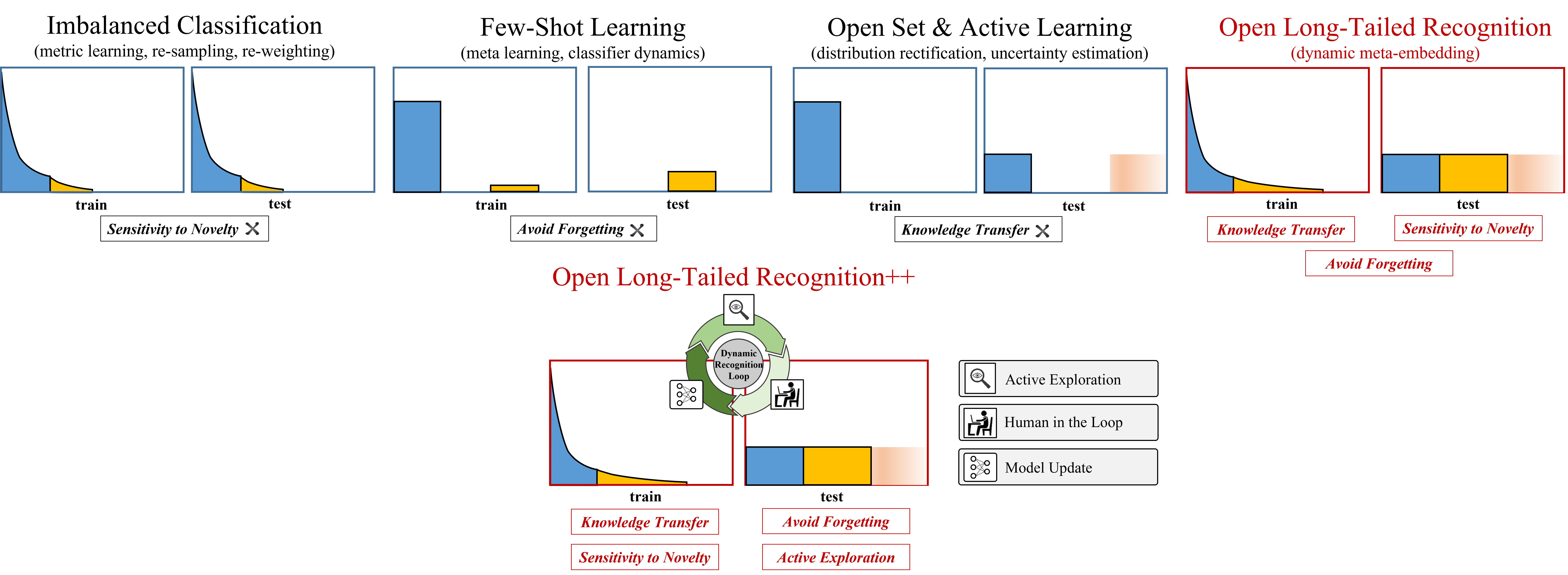}
   \caption{
   \textbf{Comparison between our proposed OLTR++ and related existing tasks.} Our task, in a more realistic setting, is a combination of the three existing tasks (imbalanced classification, few-shot learning, and open-set recognition) in addition to active exploration, which constitutes a dynamic learning loop that can facilitate robust deployment of vision systems.}
  \label{fig:relatedwork}
\end{figure*}

Although natural data distributions contain head, tail, and open classes, existing classification approaches focus mostly on either the head~\cite{deng2009imagenet, lin2014microsoft} or the tail~\cite{vinyals2016matching, lake2015human}, and often in a closed setting~\cite{wang2017learning, miao2019insights} (Fig.~\ref{fig:relatedwork}). We thus formally study {\it Open Long-Tailed Recognition++} (OLTR++) arising from natural data settings. A practical system should be able to work for a few head and many tail categories, to generalize the concept of a single category from only a few known instances, as well as to acknowledge and explore novelty upon an instance of an unseen or open category. We define OLTR++ as learning from long-tail and open-end distributed data and evaluating the classification accuracy over a balanced test set which includes head, tail, and open classes in a continuous spectrum (Fig. \ref{fig:intro}).

% Traditional deep learning  models are good at capturing the big data of head classes ~\cite{krizhevsky2012imagenet, he2016deep}; more recently, few-shot learning methods have been developed for the small data of tail classes ~\cite{snell2017prototypical, hariharan2017low}. 

%In this work, we advocate an open long-tailed recognition (OLTR) setting in order to imitate what our human vision systems encounter in the daily life. In particular, the training set consists of the classes whose frequency distribution is long-tailed and the test set is contaminated by an open-set of unlabeled data which do not belong to any of the classes of interest. Our objective is to train a robust classifier using such a training set. It will be tested against novel instances of any classes --- the classifier outputs either a class label or an ``open-set'' label for an input example. 
% This OLTR setting evaluates visual recognition systems in a more comprehensive and more realistic manner than the closed setting. This OLTR setting serves a more comprehensive and more realistic touchstone for evaluating visual recognition systems than the previous settings. 

% \begin{table*}[t]
%     \centering
%     \resizebox{1.0\textwidth}{!}{%
%     \begin{tabular}{l|c|c|c|c|c}
%     \Xhline{1pt}
%     {\bf Task Setting} & {\bf Imbalanced Train/Base Set} &  {\bf \#Instances in Tail Class } & {\bf Balanced Test Set} & {\bf Open Class} & {\bf Evaluation: Accuracy Over ?} \\ \hline \hline
%     Imbalanced Classification & $\checkmark$ & 20$\sim$50  & $\times$ & $\times$ &  all classes \\ \hline
%     Few-Shot Learning & $\times$ & 1$\sim$20 & $\checkmark$ & $\times$ &   novel classes \\ \hline
%     Open-Set Recognition & $\times$ & N/A & $\checkmark$ & $\checkmark$ & all classes \\ \hline
%     \bf Open Long-Tailed Recognition & $\checkmark$ & 1$\sim$20  & $\checkmark$ & $\checkmark$ & all classes \\
%     \Xhline{1pt}
%     \end{tabular}}
%     \caption{Comparison between our proposed OLTR task and related existing tasks.} \label{tab:comparison}
% \end{table*}

% OLTR++ is designed to handle not only imbalanced classification and few-shot learning in the closed world, but also open-set recognition and active learning with one integrated algorithm (Tab.~\ref{fig:relatedwork}).
% Existing classification approaches tend to focus on one aspect and deliver poorly over the entire class spectrum.

%Several challenges arise when long-tailed recognition is compounded with the open-set scenario:
% On the one hand, the tail classes should be compensated by transferring knowledge from head classes.
% On the other hand, the tail classes should be distinguished from the open classes.
%(1) the tail class representations should be compensated by transferring knowledge from head classes;
%(2) the tail classes should also be distinguished from the open classes.

The key challenges for OLTR++ are tail recognition robustness and open-set sensitivity.
As the number of training instances drops from thousands in the head class to a few in the tail class, we should prevent the performance from dropping drastically. Meanwhile, for open classes, the recognition performance relies on the sensitivity to distinguish unknown samples from known classes, as well as to select informative samples for data-efficient active exploration and future model updates.

%In our algorithm, the robustness of recognition is controlled by visual knowledge sharing between head and tail classes, and the recognition sensitivity is controlled by reducing the confusion between known and open classes. \textcolor{red}{(or you need to emphasize why there is a confusion between tail and open instead of open vs head.)}
 
%Two seemingly contradictory aspects of recognition robustness and recognition sensitivity on a continuous category spectrum are addressed by OLTR++ simultaneously.

In our algorithm (OLTR++), there is a learned metric that respects both the closed-world classification and acknowledge the novelty of the open world. This metric maps images to an embedding space ({\it dynamic meta-embedding}), where visual concepts can relate to each other to improve both the tail recognition robustness and open-set sensitivity.

% OLTR++ maps images to a feature space where visual concepts can relate to each other based on a learned metric  that respects the closed-world classification and acknowledges the novelty of the open world simultaneously. 

Specifically, our {\it dynamic meta-embedding} is a combination of two components: direct feature and induced feature. 
{\bf 1)} 
Direct feature is a standard embedding that gets updated from the training data by stochastic gradient descent over the classification loss. It is usually less generalized in tail classes, compared to head classes, because of the lack of sufficient supervision.
{\bf 2)} 
Memory feature, instead, is an induced feature through a visual memory association mechanism, inspired by meta-learning methods~\cite{vinyals2016matching, duan2016rl, ba2016using}, that augments the direct feature of each image for better distinguishability. The association is learned to retrieve a summary of memory activations from the direct features of each input and then combined with the original direct features to be the meta-embeddings. This memory feature augmentation is particularly effective on tail classes for the lack of supervision to provide generalized features.

% The visual memory holds discriminative class centroids of the direct features. 
% We learn to retrieve a summary of memory activations from the direct feature, combined into a meta-embedding that is enriched particularly for the tail class.

%we propose the memory feature to relate more visual concepts to the tail classes. 
% \BG{This is where Stella raised questions because it is similar in spirit to some existing works. Shall we add some references? It is fine to acknowledge the similarity to them.}

We address open-class sensitivity by dynamically calibrating the meta-embedding with respect to the visual memory. In other words, the embedding is scaled inversely by its distance to the nearest class centroid (i.e., class memory): the farther away it is from the memory, the more likely it is an open set instance.
The distance between the embedding and the nearest centroid is then transformed into ``sample informativeness'' using an energy-based model~\cite{lecun2006tutorial}, which is further employed to select informative samples for active learning.
% \BG{This is really new to me. Perhaps elaborate this reachability a little and consider to analyze it more in the exp.}

We also adopt \emph{modulated attention}~\cite{wang2017non} to encourage the head and tail classes to use different sets of spatial features. As our meta-embedding relates head and tail classes,  our modulated attention maintains discrimination between them. 

We make the following major contributions.
%\begin{enumerate}\setlength{\itemsep}{0in}
%\item 
{\bf 1)}
We formally define the OLTR++ task, which learns from long-tail and open-end distributed data and optimizes for the overall accuracy over a balanced test set.  It provides a comprehensive and unbiased evaluation of visual recognition algorithms in practical settings.
%\item
{\bf 2)}
We propose an integrated OLTR++ algorithm with dynamic meta-embedding.  It handles tail recognition robustness by relating visual concepts among head and tail embeddings, and it handles open-class sensitivity by dynamically calibrating the embedding with respect to the visual memory. 
%\item 
{\bf 3)}
We further incorporate an energy-based model into our dynamic meta-embedding for data-efficient active learning (with only sparse human annotations), which is well suited for the constantly-changing visual world.
%\item 
{\bf 4)}
We curated three large OLTR++ datasets according to a long-tailed distribution from existing representative datasets: object-centric ImageNet~\cite{deng2009imagenet}, scene-centric Places~\cite{zhou2018places}, and face-centric MS1M datasets~\cite{guo2016ms}. We also set up benchmarks for proper OLTR performance evaluation. 
%\item 
{\bf 5)}
Our extensive experimentation on these OLTR++ datasets (as well as standardized benchmarks such as CIFAR-LT 100/10 and iNaturalist-18~\cite{van2018inaturalist}) demonstrates that our method consistently outperforms the state-of-the-art methods.  
%\end{enumerate}

The aim of this work is to advocate a new learning paradigm that can perceive and update in a dynamic world, \ie simultaneously recognizing real-world long-tailed data while actively exploring novel data with human in the loop. It is a crucial step towards embodied intelligence as well as the robust deployment of vision systems. Besides exhibiting competitive performance on long-tailed recognition, our approach also demonstrates compelling results on open class detection and active exploration with a unified framework centered around visual memory.

Our code, datasets, and models are publicly available at 
\url{https://liuziwei7.github.io/projects/LongTail.html}.
Our work fills the void in practical benchmarks for imbalanced classification, few-shot learning, open-set recognition, and active learning, enabling future research that is directly transferable to real-world applications.

%% file: sections/relatedwork.tex
\input{figures/LT_methods.tex}

%------------------------------------------------------------------------

\section{Related Works}

%Though the open long-tail learning problem~\cite{salakhutdinov2011learning, zhu2014capturing, bengio2015the, zhu2016we, ouyang2016factors, van2017devil, cui2018large} has not been formally defined, it has been tackled from different facets for decades. 

While OLTR++ has not been defined in the previous literature, there are four closely related tasks which are often studied in isolation: imbalanced classification, few-shot learning, open-set recognition, and active learning. Fig.~\ref{fig:relatedwork} summarizes their connections and key differences.

\vspace{2pt}
\noindent
\textbf{Imbalanced \& Long-Tailed Recognition.}
%
% Just like long-tail learning, imbalanced classification also studies the problem of imbalanced training samples within classes.
Imbalanced classification is an extensively studied area ~\cite{salakhutdinov2011learning, zhu2014capturing, bengio2015the, liu2015deep, zhu2016we, ouyang2016factors, liu2016deepfashion, van2017devil, cui2018large}.
While classical methods include under-sampling head classes, over-sampling tail classes, and data instance re-weighting, some recent methods apply \emph{metric learning}~\cite{huang2016learning, oh2016deep}, \emph{hard negative mining}~\cite{dong2017class, lin2017focal}, and \emph{meta learning}~\cite{ha2016hypernetworks, wang2017learning}. For example, lifted structure loss~\cite{oh2016deep} introduces margins between many training instances.  The range loss~\cite{zhang2017range} enforces data in the same class to be close and those in different classes to be far apart.
Focal loss~\cite{lin2017focal} induces an online version of hard negative mining.
And metaModelNet~\cite{wang2017learning} learns a meta regression net from head classes and uses it to construct the classifier for tail classes.
% Here, our long-tail learning considers a more extreme case than the traditional imbalanced classification. Our tail classes only contain 1$\sim$20 samples while the tail classes in imbalanced classification still have 20$\sim$50 samples. More importantly, the open classes are considered in our evaluation.

As a step forward from imbalanced classification, long-tailed recognition (where the training datasets are more imbalanced)~\cite{liu2019large, cui2019class, cao2019learning, kang2019decoupling, ye2020identifying, menon2020logit, wu2020distribution, wu2021adversarial, ren2022balanced} has attracted extensive research interests recently. In Table~\ref{tab:long_tail}, we provide a systematic overview of representative long-tailed recognition works (including those published after ours).

In our approach, we have a dynamic meta-embedding that combines the strengths of both metric learning and meta learning. On the one hand, our direct feature is updated to ensure centroids of different classes are separated from each other; On the other hand, our memory feature is generated on-the-fly in a meta learning fashion to effectively transfer knowledge from head classes to tail classes.

% \vspace{2pt}
% \noindent
% \textbf{Long-Tailed Recognition.}
%

\begin{table}[t]
    \footnotesize
    \centering
    \begin{tabular}{l|c|c}
    \Xhline{1pt}
    ~{\bf Problem}~ & ~{\bf Known Classes}~ & ~{\bf Unknown Classes}~ \\ \hline \hline
    Active Learning & \checkmark~(informativeness) &  \\ \hline
    Active Exploration & \checkmark~(informativeness) & \checkmark~(info. \& openness) \\
    \Xhline{1pt}
    \end{tabular}
    \caption{
    \textbf{Key differences between active learning and active exploration.} ``info.'' stands for informativeness.}
    \label{tab:active}
\end{table}

\vspace{2pt}
\noindent
\textbf{Few-Shot Learning.}
Few-shot learning is often addressed by meta-learning techniques~\cite{schmidhuber1993neural, bertinetto2016learning, ravi2016optimization, santoro2016meta, finn2017model, yang2018learning}.
For example, matching Network~\cite{vinyals2016matching} learns a transferable feature matching metric to go beyond given classes. And Prototypical Network~\cite{snell2017prototypical} maintains a set of separable class templates. 
Additionally, feature hallucination~\cite{hariharan2017low} and augmentation~\cite{wang2018low} have also shown effectiveness.
Since these methods focus on fast learning from novel and unseen classes, they often suffer a ``catastrophic forgetting'' for training classes
% \textcolor{red}{(need reference)}
. Few-shot learning without forgetting~\cite{gidaris2018dynamic} and incremental few-shot learning~\cite{ren2018incremental} attempt to remedy this issue by leveraging the duality between features and classifiers' weights~\cite{qiao2018few, qi2018low}. However, these methods rely on balanced training sets, while OLTR++ learns from naturally long-tailed training sets instead.

Our approach is closely related to meta learning with associative memory~\cite{hinton1987using, schmidhuber1992learning, vinyals2016matching, duan2016rl, ba2016using, munkhdalai2017meta}. Compared to prior arts, our memory feature has two advantages: {\bf 1)} it transfers knowledge to both head and tail classes adaptively via a learned concept selector; {\bf 2)} it is fully integrated into the network without episodic training, thus suitable for large-scale applications.  

% Here we design ``memory feature'' for knowledge transfer in long-tailed scenario.

\vspace{2pt}
\noindent
\textbf{Open-Set Recognition.}
Open-set recognition~\cite{scheirer2013toward, bendale2016towards}, or out-of-distribution detection~\cite{devries2018learning, liang2017enhancing, liu2020energy}, aims to re-calibrate the sample confidence in the presence of open classes. 
One of the representative techniques is OpenMax~\cite{bendale2016towards}, which fits a Weibull distribution to model logits.
However, when there are both open and tail classes, distribution-fitting based methods often confuse the two because of the less generalized features of tail classes.
% To accommodate for the open long-tailed setting in this work, 
%
Instead of calibrating the output logits, our OLTR++ approach incorporates a confidence estimation mechanism into feature learning and dynamically re-scale the meta-embedding \wrt the learned visual memory, such that samples from known classes are expected to be closer to the memory compared to novel samples.

\vspace{2pt}
\noindent
\textbf{Open-World Recognition.}
Open-world recognition~\cite{bendale2015towards, boult2019learning, joseph2021towards} is a closely related field whose goal is to distinguish ``unknown unknown classes'' from ``known known classes''. \cite{bendale2015towards} also considers a dynamic setting where unknown classes are continuously added and detected, and examines the influence of unknown classes on the accuracy of known classes.
\cite{joseph2021towards} further incrementally learns the new classes. Once they are detected as unknown and an oracle provides labels for the objects of interest among all the unknowns.
Here we advocate this dynamic-world endeavor: instead of just detecting the unknown classes, we aim to recognize the semantic label of the unknown classes.

\vspace{2pt}
\noindent
\textbf{Zero-Shot Learning.}
Zero-shot learning (ZSL)~\cite{lampert2013attribute,xian2018zero,changpinyo2016synthesized} is also a promising direction for recognizing novel classes. ZSL aims to learn the association between base and novel class features with the aid of certain shared semantic knowledge (\eg attributes, word2vec), which is not directly applicable here. In comparison, our active exploration is more focused on the annotation-efficiency of recognizing novel classes, \ie using less human annotations to achieve acceptable accuracies on novel classes.

\vspace{2pt}
\noindent
\textbf{Active Learning.}
Active learning aims to explore unlabeled data with an oracle annotator that provides ground truth labels to a few selected samples. The central issue here is the exploration efficiency, \ie obtaining higher performance with less oracle queries. The representative works can be roughly categorized into two realms: generation-based methods~\cite{zhu2017generative, mayer2020adversarial} and selection-based methods~\cite{gal2017deep, sinha2019variational}. 
% Here we focus on the latter direction. 
% \textcolor{red}{(Difference?)}

However, existing active learning methods mainly work in closed-world setting, where they focus on selecting informative samples for the known categories to improve the performance.
Here we study a realistic yet more challenging problem, selecting informative samples from a mixture of known and unknown categories so as to recognize both of them, which we coined as ``active exploration''. Their key differences are listed in Table~\ref{tab:active}. Our active exploration considers both informativeness and openness during sample selection.

\begin{table}[t]
    \footnotesize
    \centering
    \begin{tabular}{l|c|c}
    \Xhline{1pt}
    ~{\bf Problem}~ & ~{\bf Imbalanced Asp.}~ & ~{\bf Optimization Obj.}~ \\ \hline \hline
    Fairness Analysis & sensitive attributes  & attribute-wise criteria \\ \hline
    Long-Tailed Recognition & categories  & acc. on all categories \\
    \Xhline{1pt}
    \end{tabular}
    \caption{
    \textbf{Key differences between fairness analysis and open long-tailed recognition.} ``asp.'' stands for aspects while ``obj.'' stands for objectives.}
    \label{tab:fairness}
\end{table}

\begin{figure*}[t]
  \centering
  \includegraphics[width=1.0\textwidth]{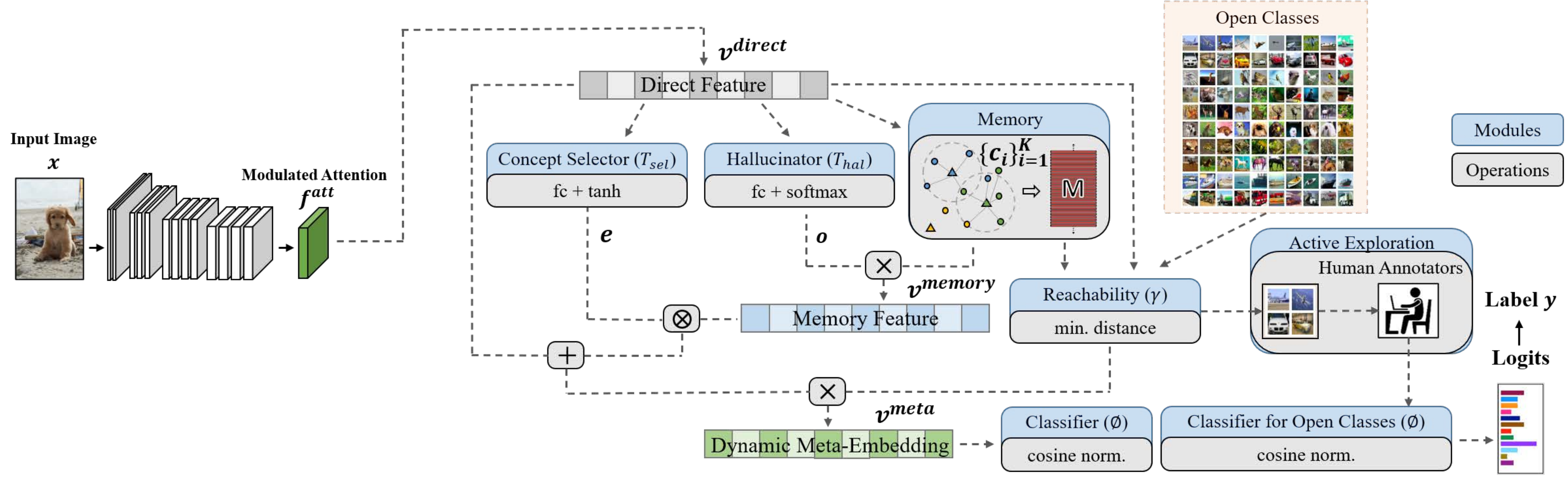}
   \caption{\textbf{OLTR++ framework overview.} There are two main modules: dynamic meta-embedding and modulated attention. The embedding relates visual concepts between head and tail classes, while the attention discriminates between them. The {\emph reachability} separates known and open classes.}
  \label{fig:pipeline}
\end{figure*}

\vspace{2pt}
\noindent
\textbf{Incremental Learning.}
Incremental learning aims to continually learn new tasks (\eg novel classes) without catastrophic forgetting. Extensive research has been performed from different perspectives: neural architectures~\cite{rusu2016progressive,lee2020neural}, replay/rehearsal mechanisms~\cite{rebuffi2017icarl,aljundi2019gradient}, parameter regularization~\cite{kirkpatrick2017overcoming,zenke2017continual} and learning techniques~\cite{li2017learning,ebrahimi2020adversarial}. Here our dynamic learning loop aims to achieve continual recognition of novel classes in a data-efficient manner.

\vspace{2pt}
\noindent
\textbf{Fairness Learning/Analysis.}
The open long-tailed recognition proposed in our work also has an intrinsic relationship to fairness analysis~\cite{dwork2012fairness, zemel2013learning, madras2018learning, mitchell2018model, anne2018women}. 
Their key differences are listed in Table~\ref{tab:fairness}.
On the problem setting side, both open long-tail recognition and fairness analysis aim to tackle the imbalance existed in real-world data.
Open long-tailed recognition focuses on the longtail-ness in both known and unknown categories while fairness analysis deals with the bias in sensitive attributes such as male/female and white/black.

On the methodology side, both open long-tailed recognition and fairness analysis aim to learn transferable representations.
Open long-tailed recognition optimizes for the overall accuracy of all categories while fairness analysis optimizes for several attribute-wise criteria.
The preliminary results in Table~\ref{tab:benchmark_megaface} demonstrates that our proposed dynamic meta-embedding is also a promising solution to fairness analysis.

% \noindent
% \textbf{Knowledge Transfer.}
% %
% Researchers have also viewed the visual long-tail learning problem from a transfer learning perspective. They attempt to tackle the problem by transferring the abundant knowledge obtained from the head classes to the tail classes via \emph{visual features, classifier weights and class distribution priors}. 
% For instance, Van Horn \etal~\cite{van2017devil} pointed out long-tail phenomenon exists in many real-world settings and proposed a progressively fine-tuning scheme, which aims to re-use the well-trained head class features to the tail classes.
% MetaModelNet~\cite{wang2017learning} took another route and transferred knowledge on the model parameter space.
% Their basic assumption is that there exists an universal weight transform that can convert a under-trained classifier to a well-trained classifier.
% Range Loss~\cite{zhang2017range} enforces the distribution of each tail class to be compact by itself as well as far from other head classes.
% In this way, the tail class features will not be overwhelmed by head classes, and thus can be discriminated during testing.
% Though promising results have been achieved, these methods usually rely on strong assumptions on classifier dynamics or distribution priors that are not readily applied to large-scale datasets.
% Our approach, on the other hand, integrates the contextual information into the learning process and enable knowledge transfer among all classes.

%% file: figures/LT_methods.tex
\begin{table*}[t]
  \centering
  \begin{threeparttable}
  \small
    \begin{tabular}{l|l|c|cccc}
    \toprule
    \toprule
    \multicolumn{1}{c|}{\textbf{Method}} & \multicolumn{1}{c|}{\textbf{Formulation}} & \multicolumn{1}{c|}{\textbf{Venue}} &
    \multicolumn{1}{c}{\textbf{CIFAR-10}} &
    \multicolumn{1}{c}{\textbf{CIFAR-100}} &
    \multicolumn{1}{c}{\textbf{ImageNet-LT}} & \multicolumn{1}{c}{\textbf{Places-LT}} \\
    \midrule
Vanilla FC~\cite{he2016deep} & $W_i^Tf(x)$ & CVPR 16 & 70.4 & 38.3 & 20.9 & 27.2 \\
% Re-sampling~\cite{shen2016relay} & $r(y) \propto 1/n_i$ & ECCV 16 & & & & \\
Hard Example mining~\cite{lin2017focal} & $r(y) = \left( 1-p_y \right)^\gamma$ & ICCV 17 & 70.4 & 38.4 & 30.5 & 34.6 \\
Re-Weighting~\cite{cui2019class} & $r(y) = (1-\beta) / (1 - \beta^n_y)$ & CVPR 19 & 74.6 & 36.0 & 29.7 & 38.9 \\
Memory \textbf{(ours)}~\cite{liu2019large} & $W_i^Tf^{memory}(x)$ & CVPR 19 & 76.3 & 41.2 & 39.6 & 39.3 \\
\midrule
Class-Aware Margin~\cite{cao2019learning} & $W_i^Tf(x) - \mathbbm{1}\{i=y\} \cdot \delta_i$ & NeurIPS 19 & 77.0 & 42.0 & 36.2 & 35.7 \\
% Class-aware temperature~\cite{ye2020identifying} & $W_i^Tf(x)\cdot (n_i / n_{max})^\gamma$ & ArXiv 20 & & & & \\
Classifier Re-Scaling~\cite{kang2019decoupling} & $s_i \cdot W_i^Tf(x)$ & ICLR 20 & - & 43.2 & 41.4 & 36.7 \\
Bilateral Branch~\cite{zhou2020bbn} & $(W_i^{bilateral})^Tf(x)$ & CVPR 20 & 79.8 & 42.6 & - & - \\
Multi Experts~\cite{wang2020long} & $(W_i^{expert})^Tf(x)$ & ICLR 21 & - & 47.0 & 54.4 & - \\
\bottomrule
\bottomrule
\end{tabular}%
% \begin{tablenotes}
% \small
% \item Suppose there are $C$ classes in total with $n_i, i \in \{1,2,..., C\}$ samples for class $i$. We denote $f(x)$ as the deep feature extracted from image $x$ and $W = [W_1, ..., W_C]$ as the classifier weight vectors.
% \end{tablenotes}
\end{threeparttable}
\caption{A systematic overview of representative long-tailed recognition works (including those published after ours). Suppose there are $C$ classes in total with $n_i, i \in \{1,2,..., C\}$ samples for class $i$. We denote $f(x)$ as the deep feature extracted from image $x$ and $W = [W_1, ..., W_C]$ as the classifier weight vectors. The accuracies are reproduced with their released code.}
\label{tab:long_tail}%
\end{table*}

%% file: sections/approach.tex
%------------------------------------------------------------------------
\section{Our OLTR++ Model}

We propose to map an image to a feature space such that visual concepts can relate to each other based on a learned metric that respects the closed-world classification while acknowledging the novelty of the open and dynamic world. Our model has three major modules (Fig.~\ref{fig:pipeline}): \emph{dynamic meta-embedding}, \emph{modulated attention}, and \emph{active exploration}. The first relates and transfers knowledge between head and tail classes, and the last two maintain discrimination between them with human-in-the-loop. 

% Figure~\ref{fig:pipeline} depicts the overall work flow
% \footnote{Notations in the figure will be explained in the text. Please also see the {supplementary materials} for an aggregated description of them.} 

 %Moreover, we also devise for the dynamic meta-embedding a reachability weight which prevents the open-set --- images of not our interest --- from reaching the meta memory. 

% In what follows, we describe the two in detail. 
%Below we describe the two in detail.

%Below we explain each module in details\footnote{Please refer to \textbf{supplementary material} for the notation summary.}: 

%------------------------------------------------------------------------
\subsection{Intuitive Explanation of Our Approach}
\label{sec:explanation}

\begin{figure}[t]
  \centering
  \includegraphics[width=0.45\textwidth]{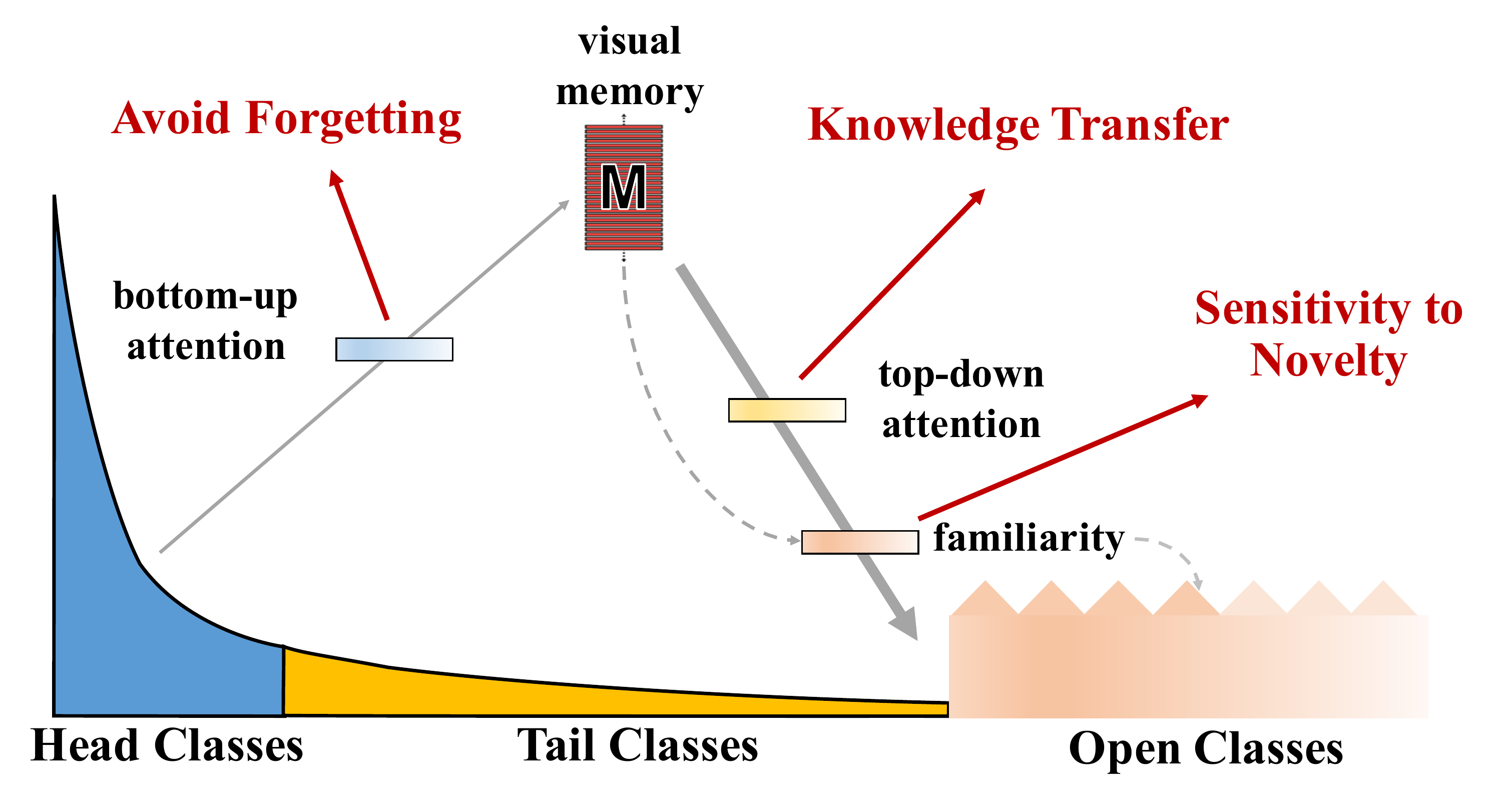}
  \caption{\textbf{Intuition explanation} of our approach.}
  \label{fig:intuition}
\end{figure}

In this section, we give an intuitive explanation of our approach that tackles the problem open long-tailed recognition.
From the perspective of knowledge gained from observation (\ie training set), head classes, tail classes and open classes form a continuous spectrum as illustrated in Fig.~\ref{fig:intuition}.

Firstly, in our approach, we obtain a \emph{visual memory} by aggregating the knowledge from both head and tail classes.
Then the visual concepts stored in the memory are infused back as associated ``fast feature'' to enhance the original ``slow feature''.
In other words, we use induced knowledge (\ie ``fast feature'') to assist the direct observation (\ie ``slow feature'').
We further learn a \emph{concept selector} to control the amount and type of infused ``fast feature''.
Since head classes already have abundant direct observation, only a small amount of ``fast feature'' is needed.
On the contrary, tail classes suffer from scarce observation, the associated visual concepts in ``fast feature'' are more beneficial to tail classes than to head classes.
Finally, we calibrate the confidence of open classes by calculating their \emph{reachabilities} (\ie feature space distances) to the obtained visual memory (i.e., class centroids).
All together, we provide a comprehensive treatment to the full spectrum of head, tail and open classes, improving the performance on all categories.

\subsection{Dynamic Meta-Embedding}
Dynamic meta-embedding is a combination of a direct image feature and an associated memory feature, where the feature norm indicates the familiarity to known classes.

Consider a convolutional neural network (CNN) with a softmax output layer for classification. The  second-to-the-last layer is the feature and the last layer is the linear classifier (cf.\ $\phi(\cdot)$ in Fig.~\ref{fig:pipeline}).
%We view an end-to-end neural classification network as two parts: a feature extractor $v_n:=F(x_n)$, where $x_n$ is the input image, and a linear classifier $y_n:=H(v_n)$. 
%Given an input image $x_{n}$ to be classified into its category label $y_{n}$, a conventional practice~\cite{krizhevsky2012imagenet} would be: (1) using feature extractor to obtain visual feature\footnote{We omit pooling here for simplicity.}: $v_{n} = F(x_{n})$; (2) training a classifier to get the final prediction: $y_{n} = H(v_{n})$.
The feature and the classifier are jointly trained in an end-to-end fashion.  Let $v^{direct}$ denote the {\it direct feature} extracted from an input image.  The final classification accuracy largely depends on the quality of this direct feature. 
% (We summarize the notations used in the paper in Table~\ref{tab:notation}.)

% \begin{table}[t]
%     \small
%     \centering
%     \begin{tabular}{l|c}
%     \Xhline{1pt}
%     ~{\bf Notation}~ & ~{\bf Meaning}~  \\ \hline \hline
%     $x$ & input image \\ \hline
%     $y$ & category label \\ \hline
%     $f$ & the original feature map \\ \hline
%     $f^{att}$ & feature map after modulated attention \\ \hline
%     $F(\cdot)$ & feature extractor \\ \hline
%     $\phi(\cdot)$ & classifier \\ \hline
%     $c_{i}$ & discriminative centroid \\ \hline
%     $G$ & local graph \\ \hline
%     $M$ & visual memory \\ \hline
%     $v^{slow}$ & slow feature \\ \hline
%     $v^{fast}$ & fast feature \\ \hline
%     $o$ & hallucinated coefficients from visual memory \\ \hline
%     $e$ & concept selector \\ \hline
%     $\gamma$ & confidence calibrator \\ \hline
%     $v^{meta}$ & dynamic meta-embedding \\ 
%     \Xhline{1pt}
%     \end{tabular}
%     \vspace{-8pt}
%     \caption{\textbf{Summary of notations.}}
%     \label{tab:notation}
% \end{table}

While a feed-forward CNN classifier works well with big training data~\cite{deng2009imagenet, krizhevsky2012imagenet}, it lacks sufficient supervised updates from small data in tail classes.  We propose to enrich direct feature $v^{direct}$ with visual concepts in a memory module through a memory feature $v^{memory}$, which is derived from a memory bank that captures visual concepts of each training classes. This mechanism is similar to the memory components in meta learning~\cite{santoro2016meta, munkhdalai2017meta}. We denote the resulting feature  {\it meta embedding} $v^{meta}$. 
%To compensate for the information gap required for a successful prediction, in this work we propose \emph{dynamic meta-embedding} $v_{n}^{meta}$ that enhances the ``direct feature'' $v_{n}^{direct}$ with a ``memory feature'' $v_{n}^{memory}$, which encodes the related visual concepts by associating a memory module $M$. By dynamically transferring visual knowledge from ``memory feature'' to ``direct feature'', \emph{dynamic meta-embedding} evolves to be an effective and robust embedding in this open long-tail setting.
%Both our memory feature $v^{memory}$ and  meta-embedding $v^{meta}$ depend on direct feature $v^{direct}$.

%\subsubsection{Generating Memory Feature}

\begin{figure}[t]
  \centering
  \includegraphics[width=0.45\textwidth]{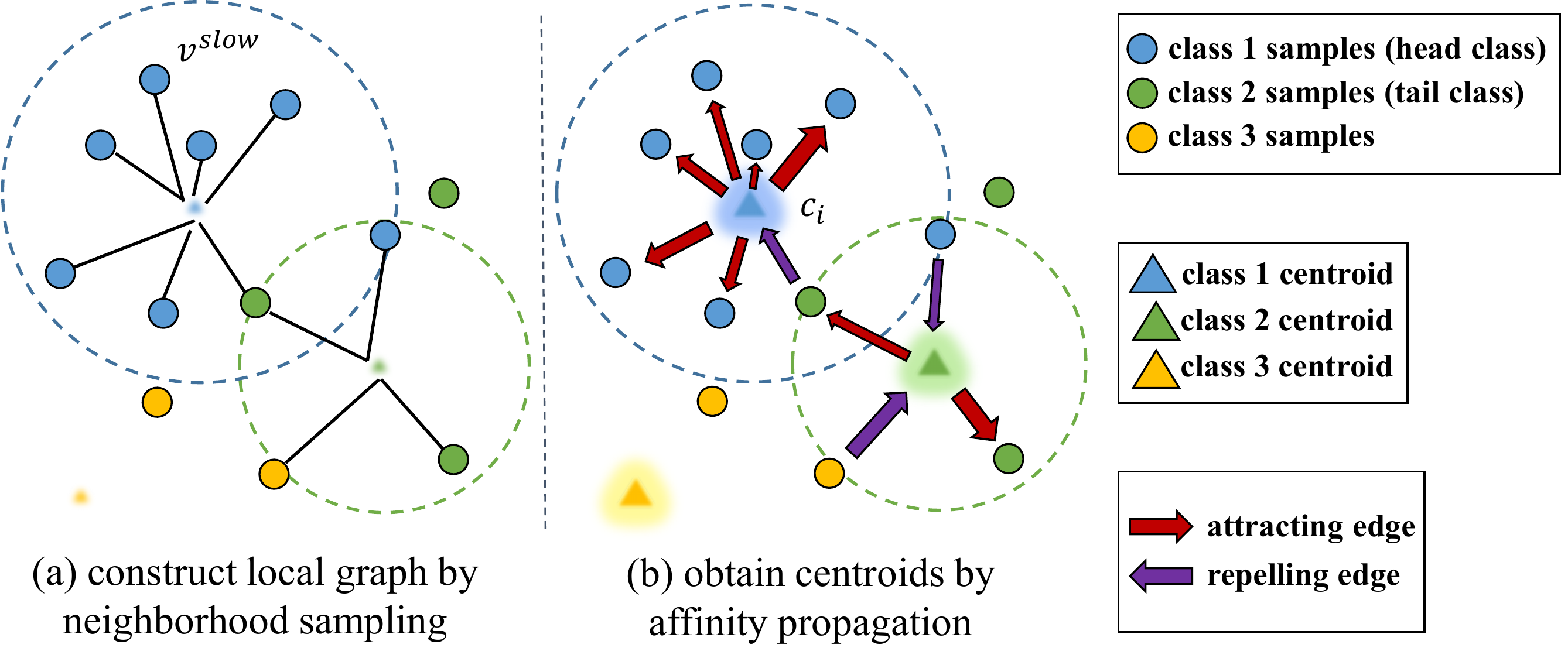}
  \caption{\textbf{The discriminative centroids} constitute our visual memory, which are obtained with two iterative steps, neighborhood sampling and affinity propagation.}
  \label{fig:centroid}
\end{figure}

% Unlike the direct feature, the memory feature captures visual concepts from training classes, retrieved from a memory with a much shallower model.

% \begin{figure}[t]
%   \centering
%   \includegraphics[width=0.45\textwidth]{figures/centroid.pdf}
%   \caption{\textbf{The discriminative centroids constitute our visual memory}, which are obtained with two iterative steps, neighborhood sampling and affinity propagation.}
%   \label{fig:centroid}
% \end{figure}

\vspace{2pt}
\noindent
\textbf{Learning Visual Memory $M$.}
To construct the memory bank, we follow~\cite{hsu2017learning,deng2021joint} on class structure analysis and adopt discriminative centroids as the basic building block.  Let $M$ denote the visual memory of all the training data, $M=\{c_{i}\}_{i=1}^{K}$ where $K$ is the number of training classes.
Compared to alternatives ~\cite{wen2016discriminative, snell2017prototypical}, this memory is appealing for our OLTR++ task:  it is almost effortlessly and jointly learned alongside the direct features $\{v_{n}^{direct}\}$, and it considers both intra-class compactness and inter-class discriminativeness.  
% More concretely, as illustrated in Fig.~\ref{fig:centroid}, the discriminative centroids are obtained through the following steps.

More concretely, as illustrated in Fig.~\ref{fig:centroid}, we compute class centroids in two steps.
\begin{enumerate}[leftmargin=*]
% {\bf 1)} 
\item \textbf{Neighborhood Sampling:} We sample both intra-class and inter-class examples to compose a mini-batch during training. These examples are grouped by their class labels and the centroid $c_i$ of each group is updated by the direct feature of this mini-batch, which can be formulated as:
\begin{align}
\Delta c_{i} = \frac{\sum_{b=1}^{B}\mathbbm{1}(y_{b}=i)\cdot(c_{i}-x_{b})}{1+\sum_{b=1}^{B}\mathbbm{1}(y_{b}=i)},
\end{align}
where $\mathbbm{1}(\cdot)$ is the indicator function, and $B$ is the batch size.
% Then the local graph $G$ is constructed on feature space with both ``direct feature'' $v_{n}^{direct}$ and the initialized discriminative centroids $\{c_{i}\}_{i=1}^{K}$.   
% {\bf 2)} 
\item \textbf{Affinity Propagation:} We alternatively update the direct feature $v^{direct}$ and the centroids to minimize the distance between each direct feature and the corresponding class centroids and maximize the distance to other centroids.  
% The centroids are initialized as  the means of the classes with roughly calculated direct features. %More details are included in the supplementary materials.
Note that the ``repelling edges'' in Fig.~\ref{fig:centroid} are calculated through a large margin loss $L_{LM}$ as described in Eqn.~\ref{eqn:largemargin}.
\end{enumerate}
At the end of the training, we obtain a visual memory module $M$ containing important visual concepts within the dataset.
% The step-by-step procedure for obtaining discriminative centroids $\{c_{i}\}_{i=1}^{K}$ is further illustrated in Fig.~\ref{fig:centroid}.

\vspace{2pt}
\noindent
\textbf{Composing Memory Feature $v^{memory}$.}
For an input image, $v^{memory}$ enhances its direct feature when training data are limited (as in the tail class). The memory feature relates class centroids in the memory to transfer knowledge to tail classes:
\begin{align}
v^{memory} = o^T M := \sum_{i=1}^K o_{i}c_i, 
\end{align}
where $o\in\mathbb{R}^K$ is the coefficients hallucinated from the direct feature. We use a lightweight neural network to obtain the coefficients from the direct feature, $o = T_{hal}(v^{direct})$.
% and defer the details of this halluciation network to the supplementary materials.

%This meta hallucinator takes the form of a one-hidden-layer MLP which associates ``direct feature'' with visual memory. 

\begin{figure}[t]
  \centering
  \includegraphics[width=0.45\textwidth]{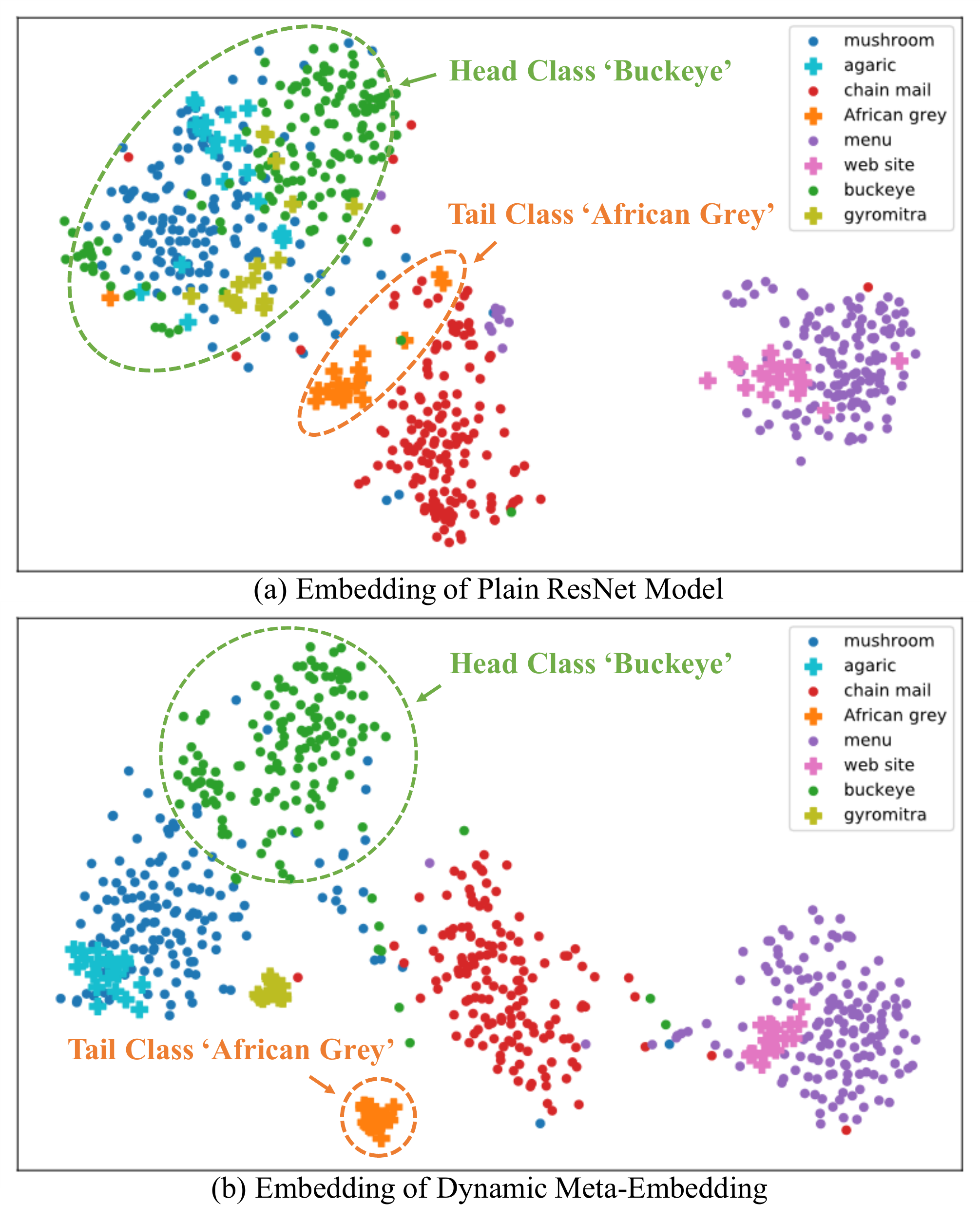}
  \caption{\textbf{t-SNE feature visualization} of (a) plain ResNet model (b) our \emph{dynamic meta-embedding}. Ours is more compact for both head and tail classes. }
  \label{fig:tsne}
\end{figure}

%\subsubsection{Obtaining Dynamic Meta-Embedding}

\vspace{2pt}
\noindent
{\bf Obtaining Dynamic Meta-Embedding}.
 $v^{meta}$  combines the direct feature and the memory feature, and is fed to the classifier for the final class prediction (Fig.~\ref{fig:tsne}):
\begin{equation}
v^{meta} = (1 / \gamma) \cdot (v^{direct} + e \otimes v^{memory}), \label{eq:embedding}
\end{equation}
where $\otimes$ denotes element-wise multiplication and $e$ is a concept selector.
$\gamma>0$ is a seemingly redundant scalar for the closed-world classification tasks.  However, in the OLTR++ setting, it serves an important role in differentiating the examples of the training classes from those of the open-set.
$\gamma$ measures the reachability~\cite{savinov2018episodic} of an input's direct feature $v^{direct}$ to the memory $M$ --- the minimum distance between the direct feature and the discriminative centroids:
\begin{align}
\hspace{-3mm}\gamma := \text{reachability}(v^{direct}, M)=\min_i \;\|v^{direct} - c_{i}\|_{2}.
\end{align}
When $\gamma $ is small, the  input likely belongs to a training class from which the centroids are derived, and a large reachability weight $1/\gamma$ is assigned to the resulting meta-embedding $v^{meta}$. Otherwise, the embedding is scaled down to an almost all-zero vector at the extreme, which is useful to encode open classes.

%To accommodate for the open-set setting, we dynamically re-scale the embedding with a calibrating confidence, which is calculated as the reachability~ from the current ``direct feature'' $v_{n}^{direct}$ to the learned visual memory $M$:

%We now describe the concept selector $e$ in Eq.~(\ref{eq:embedding}).  

As the direct feature is often good enough for the data-rich head classes, whereas the memory feature is more important for the data-poor tail classes. We design a concept selector, $e$, to adaptively select them in a soft manner. We learn a lightweight network $T_{sel}(\cdot)$ with a $\tanh(\cdot)$ activation function on $v^{direct}$:
\begin{equation}
e = \tanh(T_{sel}(v^{direct})).
\end{equation}

%Similarly, this meta selector takes the form of a MLP with one hidden layer which adaptively selects visual concepts. which learns to transfer visual concepts between head and tail classes. `$\otimes$' denotes element-wise multiplication.

\begin{figure}[t]
  \centering
  \includegraphics[width=0.48\textwidth]{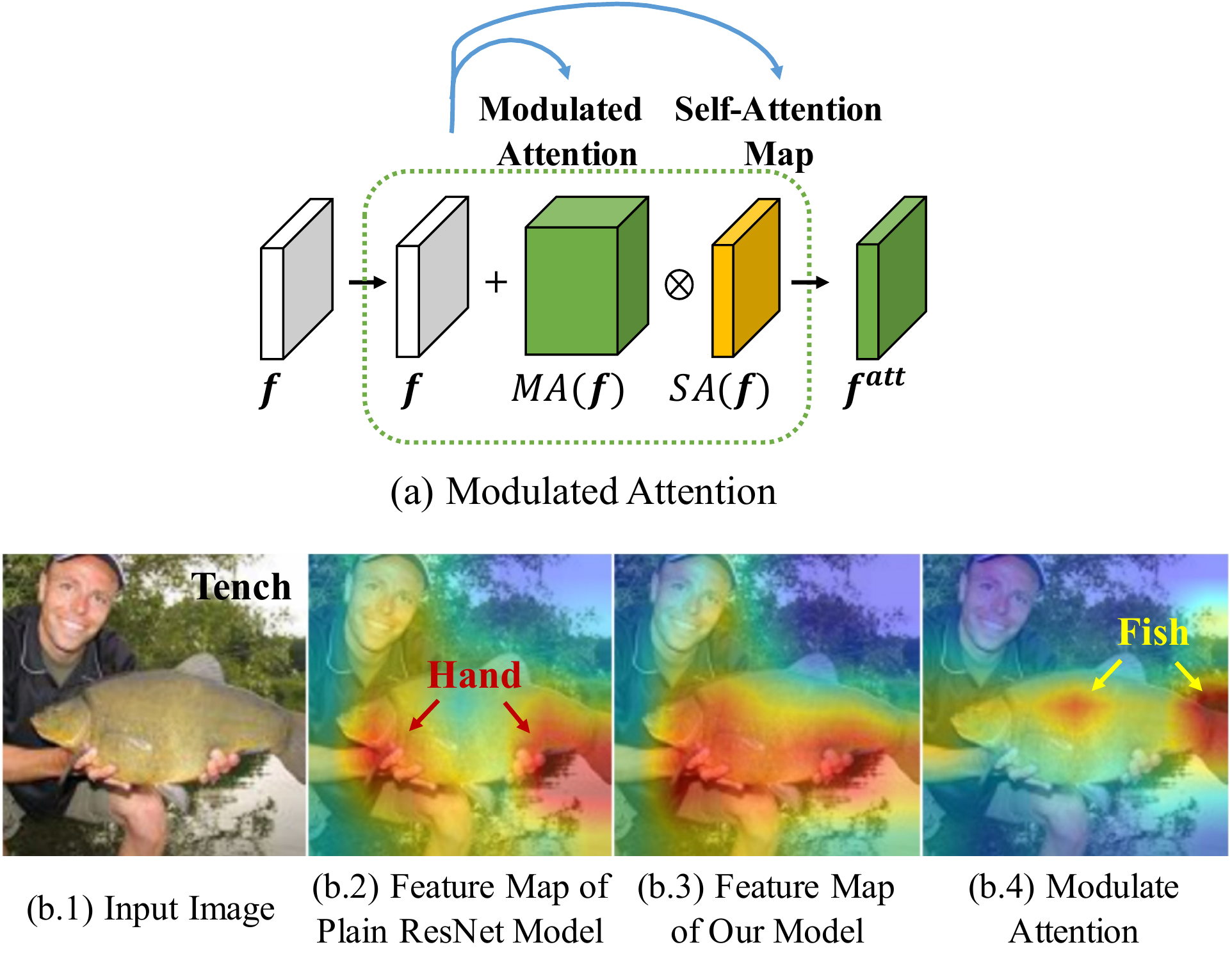}
  \caption{\textbf{Modulated attention} is spatial attention applied on self-attention maps (``attention on attention''). It encourages different classes to use different contexts, which helps maintain the discrimination between head and tail classes.}
  \label{fig:attention}
\end{figure}

\subsection{Modulated Attention}
% \BG{Better make the key techniques as clear as self-contained. Instead of referring the previous works by the self-attention and modular attention, better explain the details here.}
% While {dynamic meta-embedding} facilitates feature sharing between head and tail classes, it is also crucial to discriminate between them. 

%The direct feature $v^{direct}$, e.g., the activation at the second-to-the-last layer in ResNet~\cite{he2016deep},
% is able to fulfill this requirement to some extent. 

Since discriminative cues of head and tail classes tend to distribute at different locations in the image, we find it beneficial to enhance direct feature, $v^{direct}$, with spatial attention to separate head and tail classes. Specifically, we propose \emph{modulated attention} to encourage samples of different classes to use different contexts.
Firstly, we compute a self-attention map $SA(f)$ from the input feature map by self-correlation~\cite{wang2017non}.
It is used as contextual information and added back (through skip connections) to the original feature map. 
The \emph{modulated attention} $MA(f)$ is then designed as  conditional spatial attention applied to the self-attention map: $MA(f) \otimes SA(f)$, which allows examples to select different spatial contexts (Fig.~\ref{fig:attention}).  The final attention feature map becomes:
\begin{equation}
f^{att} = f + MA(f) \otimes SA(f),
\end{equation}
where $f$ is a feature map in CNN, $SA(\cdot)$ is the self-attention operation, and $MA(\cdot)$ is a conditional attention function~\cite{vaswani2017attention} with a softmax normalization.

Sec.~\ref{sec:ablation} shows empirically  that our attention design achieves superior performance than the common practice of applying spatial attention to the input feature map. This modulated attention (Fig.~\ref{fig:attention} (b)) could be plugged into any feature layer of a CNN. Here, we modify the last feature map only.

% \begin{figure}[t]
%   \centering
%   \includegraphics[width=0.45\textwidth]{figures/featuremap.pdf}
%   \vspace{-10pt}
% %   \caption{\small \textbf{Comparison of feature maps.} Visualization of: (a) input image, (b) feature map of plain ResNet model, (c) feature map of our model, (d) feature map of \emph{modulated attention}.}
%   \caption{\footnotesize \textbf{Comparison of feature maps.} With \emph{modulated attention}, the model can focus more on the object of interest (in this case, the tail of the tench rather than the hands of the human).}
%   \label{fig:featuremap}
%   \vspace{-8pt}
% \end{figure}

% \begin{figure*}[t]
%   \centering
%   \includegraphics[width=0.9\textwidth]{figures/ablation.pdf}
%   \vspace{-10pt}
% %   \caption{\small Ablation study of (a) the overall framework, (b) \emph{dynamic meta-embedding}, (c) \emph{modulated attention}. The performance is reported with \emph{open set} top-$1$ classification accuracy on ImageNet-LT.}
%   \caption{\footnotesize \textbf{Results of ablation study.} Dynamic meta-embedding contributes most on medium-shot and few-shot classes while modulated attention helps maintain the discrimination of many-shot classes. (The performance is reported with \emph{open-set} top-$1$ classification accuracy on ImageNet-LT.)}
%   \label{fig:ablation}
%   \vspace{-8pt}
% \end{figure*}

\begin{algorithm}[t]
    \caption{Active Exploration of Open Classes.}
    \label{alg:active}
    \begin{algorithmic}
    \INPUT{
    \\$v^{direct}$: the direct feature extracted from the open sample, 
    \\$c_{i}$: the discriminative centroid of class $i$ from visual memory, 
    \\$K$: the number of classes, 
    \\$T_{act}$: the temperature for trade-off in active exploration.}
    \For{each exploration step}
        \State Sample mini-batch $\{v^{direct}_{n}\}$.
        \State Compute the minimum distance between the direct feature and the discriminative centroid: 
        \\ \;\;\;\;\; $U_{open} \gets \min_i \;\|v^{direct} - c_{i}\|_{2}$.
        \State Compute the ratio between the first two nearest distances: 
        \\ \;\;\;\;\; $U_{info} \gets d^{sorted}_{1} \; / \; d^{sorted}_{2}$.
        \State Compute the free energy function of $v^{direct}$:
        \\ \;\;\;\;\; $E_{n} \gets -T_{act} \cdot \log \sum_{i}^{K}e^{U_{open} \cdot U_{info}/T_{act}}$.
        \State Select high-energy samples for further human annotations.
        \State Update the classifier $\phi(\cdot)$ using newly added data.
    \EndFor
    \end{algorithmic}
\end{algorithm}

\subsection{Active Exploration of Open Classes}

In the dynamic world, the model should not halt after training. We assume a continuous training, inference, annotation, and model update loop as our model actively explores the visual world over time. 
Every time our model encounters certain sample of open classes, our model will determine whether this sample is informative enough for further human annotation.  
After obtaining these human annotations in an efficient manner, our model will be continually updated according to the newly added data.

The active exploration step has three major components: 1) active sample selection based on two different types of uncertainty, 2) human-in-the-loop annotation, and 3) model update using active data annotations, all three of which constitute a dynamic recognition loop. The detailed algorithmic pipeline of our active exploration is listed in Alg.~\ref{alg:active}.

\subsubsection{Two Types of Uncertainty in Active Exploration}

Unlike the standard active learning setting that work in closed-world setting, there actually exist two types of uncertainty here in active exploration: \textbf{uncertainty in openness} and \textbf{uncertainty in informativeness}. Existing active learning algorithms are not directly applicable here since their uncertainty estimation mechanism only considers the informativeness among known classes, which is not suitable for modeling the openness between known classes and unknown classes. 

\begin{figure}[t]
  \centering
  \includegraphics[width=0.48\textwidth]{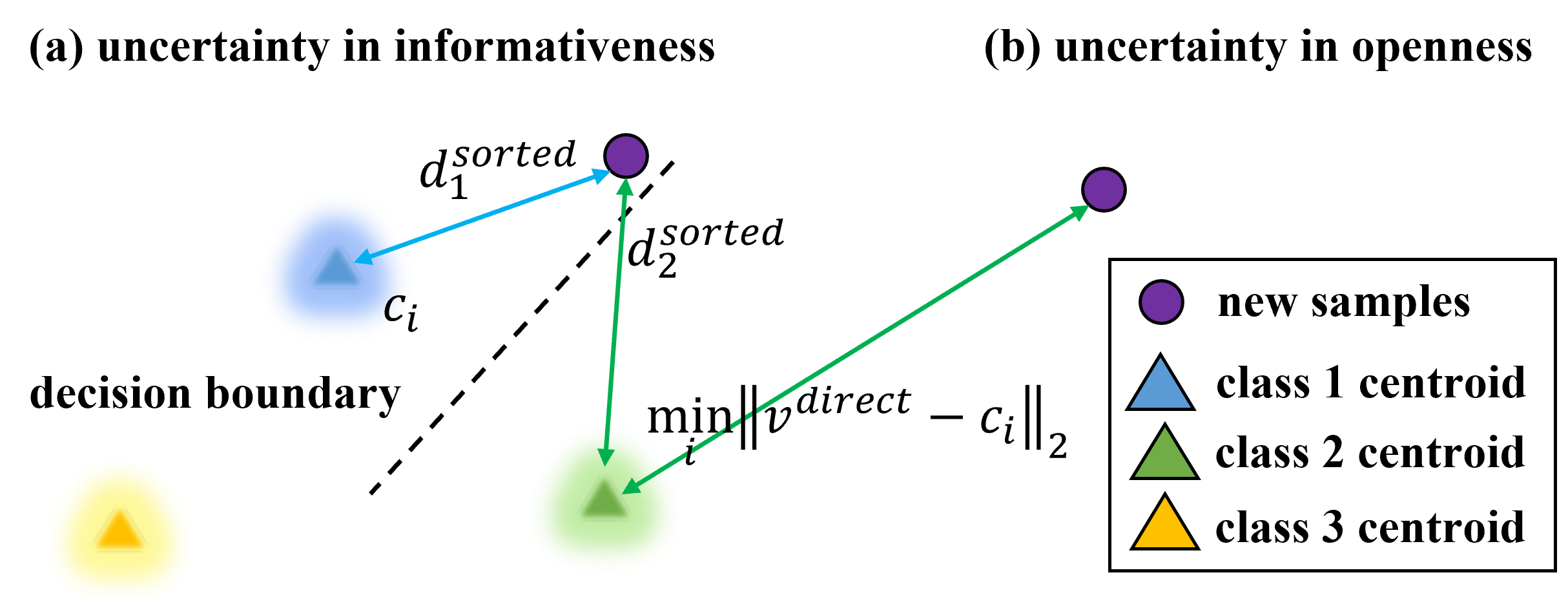}
  \caption{
  \textbf{Active exploration illustration} of (a) uncertainty in informativeness, and (b) uncertainty in openness.}
  \label{fig:uncertainty}
\end{figure}

In the following, we elaborate the modeling the two types of uncertainty in the context of active exploration:

\begin{enumerate}[leftmargin=*]
\item \textbf{Uncertainty in Openness:} We measure the openness $U_{open}$ of a new sample using the distance between its embedding and the nearest centroid, which can be formulated as:
\begin{align}
U_{open} = \min_i\|v^{direct} - c_{i}\|_{2},
\end{align}
where $v^{direct}$ is the direct feature of the new sample and $c_{i}$ is the centroid of the $i$-th class.
\item \textbf{Uncertainty in Informativeness:} Intuitively, the most informative samples would be those that lie on the decision boundaries between different classes. We first sort the distances between the embedding of the new sample to all existing class centroids in ascending order: $d^{sorted}_{n}$. Then the informativeness of a new sample is defined as the ratio between the first two nearest distances:
\begin{align}
U_{info} = d^{sorted}_{1} / d^{sorted}_{2}.
\end{align}
\end{enumerate}
These two types of uncertainty regarding new sample and class centroids are further illustrated in Fig.~\ref{fig:uncertainty}.

\begin{figure*}[t]
\centering
\includegraphics[width=0.95\textwidth]{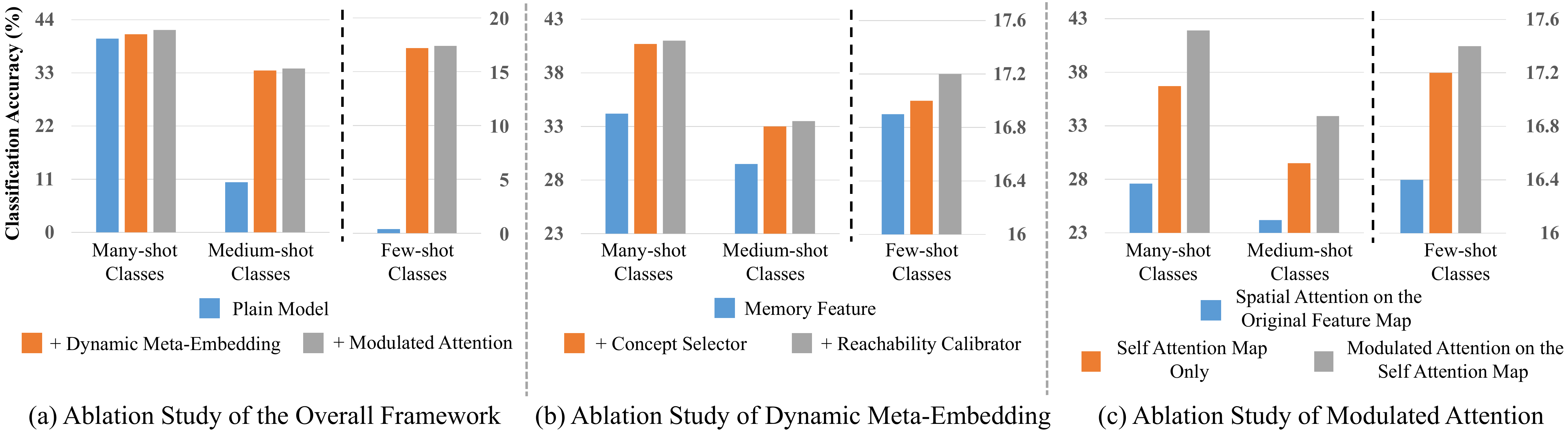}
\caption{\textbf{Results of ablation study.} Dynamic meta-embedding contributes most on medium-shot and few-shot classes while modulated attention helps maintain the discrimination of many-shot classes. The performance is reported with \emph{open-set} top-$1$ classification accuracy on ImageNet-LT.}
\label{fig:ablation}
\end{figure*}

\subsubsection{Active Sample Selection}

% At each time period when new data is encountered, the distance between the embedding and the nearest centroid is transformed into ``sample informativeness'' using the energy-based model~\cite{lecun2006tutorial}, which is further employed to select informative samples for active learning.
At each time step when new data are encountered, the selection of samples for active annotations is based on the combination of both openness and informativeness uncertainty estimation using the energy-based model~\cite{lecun2006tutorial}.
Similar to~\cite{liu2020energy}, we can express the free energy function $E(\cdot)$ of new sample $v^{direct}$ as follows:
\begin{equation}
E(v^{direct}) = -T_{act} \cdot \log \sum_{i}^{K}e^{U_{open} \cdot U_{info}/T_{act}},
\end{equation}
where $T_{act}$ is the temperature for controlling the trade-off between precision and recall in active selection.

\subsubsection{Human-in-the-Loop Annotation}

After selecting these ``open'' yet ``informative'' samples, we obtain the semantic labels of these samples by querying human annotators. In the real-world applications~\cite{miao2021iterative, zhang2022bamboo}, it can be executed through online crowdsourcing platform with quality control. Since different human annotators have different preferences for naming unknown categories, it is crucial to maintain consistency  between different samples of the same unknown category.

\subsubsection{Model Update}

Assume that we have an existing classifier $\phi_{t}(\cdot)$ at time step $t$ with weight vectors $\{w_i\}_{i=1}^K$ for $K$ known classes. And during time step $t$ to time step $t+1$, we will encounter $Z$ unknown classes. Then at time step $t+1$, our new classifier $\phi_{t+1}(\cdot)$ will be a concatenation of both known class weights and unknown class weights: $\{\{w_i\}_{i=1}^{K}, \{w_j\}_{j=K+1}^{K+Z}\}$. Both weights will be updated by the obtained active human annotations.

Note that model update is not our main contribution; there are several feasible instantiations. Since we adopt cosine classifer described in Sec.~\ref{sec:learning}, the weight of classifier and the embedding of sample can be transformed interchangeably~\cite{qi2018low,gidaris2018dynamic}. Though more sophisticated methods (\eg learning another network to generate the classifier weights for unknown classes) can be applied, here the classifier weights for unknown classes are simply hallucinated through a weighted average of the meta-embeddings from the actively selected samples for that class, where the weight is determined by $E(v^{direct})$. This classifier hallucination approach is extremely suitable for off-the-shelf deployment.

\subsubsection{Dynamic Recognition Loop}

% The aforementioned three steps can be performed iteratively to form a dynamic recognition loop. 
To accommodate for the dynamic nature of the visual world, this procedure of active sample selection, human-in-the-loop annotation, and model update repeats each time new batch of open data are encountered. Our framework maximizes learning and recognition efficiency by taking the best from both humans and machines within a synergistic collaboration, taking care of both the long-tailed and open-ended distribution existing in the natural world.

In our implementation, the feature extractor is fixed for fast adaptation, while the visual memory is updated to accommodate for the continual stream of unknown classes during the dynamic learning loop. We have further clarified in the revised paper.

% The model actively picks out low-confidence predictions for human annotation, while we accept high-confidence predictions without further human verification as accurate. These predictions are used as pseudo-labels that are included in the final data set for further model updates or downstream ecological analyses. Next, the model is updated using both pseudo-labels and the newly acquired human annotations (see Supplementary Method section for implementation details). 

% After updating the model, we evaluated model-update efficiency and sensitivity to novel categories on a validation set. Specifically, we examined: 1) the overall validation accuracy of each category after the update (i.e., update performance); 2) percentage of high-confidence predictions on validation (i.e., saved human effort for annotation); 3) accuracy of high-confidence predictions; and 4) percentage of novel categories that are detected as low-confidence predictions (i.e., sensitivity to novelty). The optimization of the algorithm is to minimize human efforts (i.e., to maximize high-confidence percentage) and to maximize model update performance and high-confidence accuracy.

\subsection{Learning}
\label{sec:learning}

\noindent
\textbf{Cosine Classifier.}
We adopt the cosine classifier~\cite{qi2018low,gidaris2018dynamic} to produce the final classification results.
Specifically, we normalize the meta-embeddings $\{v_{n}^{meta}\}$, where $n$ stands for the $n$-th input as well as the weight vectors $\{w_i\}_{i=1}^K$ of the classifier $\phi(\cdot)$ (no bias term):
\begin{equation}
\begin{split}
v_{n}^{meta} &= \frac{\|v_{n}^{meta}\|^{2}}{1+\|v_{n}^{meta}\|^{2}} \cdot \frac{v_{n}^{meta}}{\|v_{n}^{meta}\|}, \\
w_{k} &= \frac{w_{k}}{\|w_{k}\|}.
\end{split}
\end{equation}
The normalization strategy for the meta-embedding is a non-linear squashing function \cite{sabour2017dynamic}
which ensures that vectors of small magnitude are shrunk to almost zeros while vectors of big magnitude are normalized to the length slightly below $1$. This function helps amplify the effect of the reachability $\gamma$ (cf.\ Eq.~(\ref{eq:embedding})).

\vspace{2pt}
\noindent
\textbf{Loss Function.}
Since all our modules are differentiable, our model can be trained end-to-end by alternatively updating the  centroids $\{c_{i}\}_{i=1}^{K}$ and the \emph{dynamic meta-embedding} $v_{n}^{meta}$.
The final loss function $L$ is a combination of the cross-entropy classification loss $L_{CE}$ and the large-margin loss between the embeddings and the centroids $L_{LM}$:
\begin{equation}
L = \sum_{n=1}^{N} L_{CE}(v_{n}^{meta}, y_{n}) + \lambda \cdot L_{LM}(v_{n}^{meta}, \{c_{i}\}_{i=1}^{K}),
\end{equation}
% where $\lambda$ is set to $0.1$ in our experiments. 
% For more details of the losses and implementation details, please refer to our {supplementary material}.
where $\lambda$ is set to $0.1$ in our experiments 
via observing the accuracy curve on validation set.

Specifically, $L_{CE}$ is the cross-entropy loss between dynamic meta-embedding $v^{meta}_{n}$ and the ground truth category label $y_{n}$:
\begin{equation}
\begin{split}
L_{CE}(v^{meta}_{n}, y_{n}) &= y_{n}\log(\phi(v^{meta}_{n})) \\
&+ (1- y_{n})\log(1 - \phi(v^{meta}_{n})),
\end{split}
\end{equation}
where $\phi(\cdot)$ is the cosine classifier described in Eqn. 6 in the main paper.
Next we introduce the large margin loss $L_{LM}$ between the embedding $v^{meta}_{n}$ and the centroids $\{c_{i}\}_{i=1}^{K}$:
\begin{equation}
\begin{split}
L_{LM}(v^{meta}_{n}, \{c_{i}\}_{i=1}^{K}) &= \max(0, \sum_{i = y_{n}} \|v^{meta}_{n} - c_{i} \| \\
&- \sum_{i \neq y_{n}} \|v^{meta}_{n} - c_{i} \| + m),
\end{split}
\label{eqn:largemargin}
\end{equation}
where $m$ is the margin and we set it as $5.0$ in our experiments. With this formulation, we minimize the distance between each embedding and the centroid of its group and meanwhile maximize the distance between the embedding and the centroids it does not belong to.

% The learning process for \emph{modulated attention} is straightforward. 
% Below we focus on the learning of \emph{dynamic meta-embedding}, which involves the iterative updating of visual memory $M$, meta hallucinator $T_{hal}(\cdot)$ and meta selector $T_{sel}(\cdot)$ as outlined in Alg.~\ref{alg:learning}.

% \begin{algorithm}[H]
% \SetAlgoLined
% \KwResult{$\Theta$}
%  \While{not done}{
%   instructions\;
%   \eIf{condition}{
%   instructions1\;
%   instructions2\;
%   }{
%   instructions3\;
%   }
%  }
%  \caption{Dynamic Meta-Embedding Learning}
%  \label{alg:learning}
% \end{algorithm}

%% file: sections/experiments.tex
\begin{table*}[t]
\footnotesize

\begin{subtable}{\linewidth}
\centering
\bgroup
\begin{tabular}{l|cccc|cccc}
\Xhline{1pt}
\textbf{Backbone Net} & \multicolumn{4}{c|}{\footnotesize{\textbf{closed-set setting}}} & \multicolumn{4}{c}{\footnotesize{\textbf{open-set setting}}} \\
ResNet-10 & $\geqslant100$ & $<100$ \& $>20$ & $\leqslant20$ &  & $\geqslant100$ & $<100$ \& $>20$ & $\leqslant20$ & \\
\textbf{Methods} & \textbf{Many-shot} & \textbf{Medium-shot} & \textbf{Few-shot} & \textbf{Overall} & \textbf{Many-shot} & \textbf{Medium-shot} & \textbf{Few-shot} & \textbf{F-measure} \\
\hline\hline
Plain Model~\cite{he2016deep} & 40.9 & 10.7 & 0.40 & 20.9 & 40.1 & 10.4 & 0.40 & 0.295 \\
Lifted Loss~\cite{oh2016deep} & 35.8 & 30.4 & 17.9 & 30.8 & 34.8 & 29.3 & 17.4 & 0.374 \\
Focal Loss~\cite{lin2017focal} & 36.4 & 29.9 & 16.0 & 30.5 & 35.7 & 29.3 & 15.6 & 0.371 \\
Range Loss~\cite{zhang2017range} & 35.8 & 30.3 & 17.6 & 30.7 & 34.7 & 29.4 & 17.2 & 0.373 \\
~~~+~OpenMax~\cite{bendale2016towards} & - & - & - & - & 35.8 & 30.3 & 17.6 & 0.368 \\
FSLwF~\cite{gidaris2018dynamic} & 40.9 & 22.1 & 15.0 & 28.4 & 40.8 & 21.7 & 14.5 & 0.347 \\
CB Focal~\cite{cui2019class} & 35.0 & 27.9 & 21.4 & 29.7 & 34.3 & 27.4 & \textbf{21.1} & 0.361 \\
LDAM~\cite{cao2019learning} & 47.1 & 31.7 & 20.9 & 36.2 & \textbf{46.8} & 31.4 & 20.6 & 0.424 \\
Decoupling~\cite{kang2019decoupling} & - & - & - & 41.4 & - & - & - & - \\
\hline
Ours & 47.8 & 38.4 & 18.4 & 39.6 & 44.2 & \textbf{35.2} & 17.5 & \textbf{0.446} \\
\Xhline{1pt}
\rowcolor{Gray}
ResNet-50 &  &  &  &  &  &  &  & \\
\hline\hline
\rowcolor{Gray}
Plain Model~\cite{he2016deep} & 64.0 & 33.8 & 5.8 & 41.6 & - & - & - & - \\
\rowcolor{Gray}
Decoupling~\cite{kang2019decoupling} & 57.1 & 45.2 & 29.3 & 47.7 & - & - & - & - \\
\rowcolor{Gray}
RIDE~\cite{wang2020long} & \textbf{65.8} & 51.0 & 34.6 & 54.4 & - & - & - & - \\
\rowcolor{Gray}
Ours$^{\dagger}$ & 65.5 & \textbf{51.8} & \textbf{35.2} & \textbf{55.0} & - & - & - & - \\
\Xhline{1pt}
\end{tabular}
\egroup
\subcaption{ Top-$1$ classification accuracy on ImageNet-LT.}
\end{subtable}

\vspace{6pt}

\begin{subtable}{\linewidth}
\centering
\bgroup
\begin{tabular}{l|cccc|cccc}
\Xhline{1pt}
\textbf{Backbone Net} & \multicolumn{4}{c|}{\footnotesize{\textbf{closed-set setting}}} & \multicolumn{4}{c}{\footnotesize{\textbf{open-set setting}}} \\
ResNet-152 & $\geqslant100$ & $<100$ \& $>20$ & $\leqslant20$ &  & $\geqslant100$ & $<100$ \& $>20$ & $\leqslant20$ & \\
\textbf{Methods} & \textbf{Many-shot} & \textbf{Medium-shot} & \textbf{Few-shot} & \textbf{Overall} & \textbf{Many-shot} & \textbf{Medium-shot} & \textbf{Few-shot} & \textbf{F-measure} \\
\hline\hline
Plain Model~\cite{he2016deep} & \textbf{45.9} & 22.4 & 0.36 & 27.2 & \textbf{45.9} & 22.4 & 0.36 & 0.366 \\
Lifted Loss~\cite{oh2016deep} & 41.1 & 35.4 & 24.0 & 35.2 & 41.0 & 35.2 & 23.8 & 0.459 \\
Focal Loss~\cite{lin2017focal} & 41.1 & 34.8 & 22.4 & 34.6 & 41.0 & 34.8 & 22.3 & 0.453 \\
Range Loss~\cite{zhang2017range} & 41.1 & 35.4 & 23.2 & 35.1 & 41.0 & 35.3 & 23.1 & 0.457 \\
~~~+~OpenMax~\cite{bendale2016towards} & - & - & - & - & 41.1 & 35.4 & 23.2 & 0.458 \\
FSLwF~\cite{gidaris2018dynamic} & 43.9 & 29.9 & 29.5 & 34.9 & 38.1 & 19.5 & 14.8 & 0.375 \\
\hline
CB Focal~\cite{cui2019class} & 43.4 & 39.1 & \textbf{30.5} & 38.9 & 42.3 & 37.7 & \textbf{28.8} & 0.490 \\
LDAM~\cite{cao2019learning} & 45.6 & 37.8 & 23.9 & 35.7 & 45.6 & 37.7 & 23.5 & 0.485 \\
Decoupling~\cite{kang2019decoupling} & 40.6 & 39.1 & 28.6 & 37.3 & - & - & - & - \\
\hline
Ours & 44.0 & \textbf{40.6} & 28.5 & \textbf{39.3} & 43.7 & \textbf{40.2} & 28.0 & \textbf{0.500} \\
\Xhline{1pt}
\end{tabular}
\egroup
\subcaption{ Top-$1$ classification accuracy on Places-LT.}
\end{subtable}
\caption{
\textbf{Benchmarking results on (a) ImageNet-LT and (b) Places-LT.} Our approach provides a comprehensive treatment to all the many/medium/few-shot classes as well as the open classes with substantial advantages on all aspects. $^{\dagger}$ denotes using the two-headed ResNet-50 backbone following~\cite{wang2020long}.}
\label{tab:benchmark_imagenet}
\end{table*}

\begin{table}[t]
\footnotesize
\centering
\begin{tabular}{l|cc}
\Xhline{1pt}
\textbf{Method} & \textbf{FPR ($95\%$ TPR)} & \textbf{Detection Error} \\
\hline\hline
Softmax Pred.~\cite{hendrycks2016baseline} & 82.2 & 43.6 \\
Ours & 51.5 & 29.9 \\
ODIN~\cite{liang2017enhancing}$^{\dagger}$ & 44.3 & 24.6 \\
Energy OOD~\cite{liu2020energy}$^{\dagger}$ & 40.7 & 21.1 \\
Ours$^{\dagger}$ & \textbf{35.4} & \textbf{18.0} \\
\Xhline{1pt}
\end{tabular}
\caption{
\textbf{Open class detection error (\%) comparison.} It is performed on the standard open-set benchmark, CIFAR100 + TinyImageNet (resized). ``$\dagger$'' denotes the setting where open samples are used to tune algorithmic parameters.}
\label{tab:openset}
\end{table}

%------------------------------------------------------------------------
\section{Experiments}

\noindent
\textbf{Datasets.}
% \footnote{Please refer to our supplementary materials for more dataset specifications and their corresponding evaluation protocols.}
%
We curated three open long-tailed benchmarks, ImageNet-LT (object-centric), Places-LT (scene-centric), and MS1M-LT (face-centric), respectively.

\begin{enumerate}[leftmargin=*]
\item ImageNet-LT: We constructed a long-tailed version of the original ImageNet-2012~\cite{deng2009imagenet} by sampling a subset  following Pareto distribution with power value $\alpha$=6. Overall, it has $115.8$K images from $1000$ categories, with  maximally $1280$ images per class and minimally $5$ images per class. 
The additional classes of images in ImageNet-2010 are tagged as the open set.
% We randomly select $20$ images per class from the origin training set as our val set. The origin val set of ImageNet dataset is used as our test set.
\item Places-LT: It contains $184.5$K images from $365$ categories, with the maximum of $4980$ images per class and the minimum of $5$ images per class. 
The gap between the head and tail classes are even larger than ImageNet-LT. The test images from Places-Extra69 are tagged as the additional open-set.
% We also randomly select $20$ images per class from the origin training set as our validation set. The origin validation set of Places dataset is used as our testing set.
\item MS1M-LT: 
% The original MS1M-ArcFace dataset~\cite{guo2016ms, deng2018arcface} contains about 5.8M images with 85K identities. 
To create a long-tailed version of the MS1M-ArcFace dataset~\cite{guo2016ms, deng2018arcface}, we sampled images for each identity with a probability proportional to the image numbers of each identity. It has $887.5$K images and $74.5$K identities, with a long-tailed distribution.
To inspect the generalization ability of our approach, the performance is evaluated on the MegaFace benchmark~\cite{kemelmacher2016megaface}, which has no identity overlap with MS1M-ArcFace.
\end{enumerate}

\noindent
\textbf{Network Architectures.}
Following~\cite{hariharan2017low, wang2018low, gidaris2018dynamic}, we employed a ResNet-10~\cite{he2016deep} trained from scratch as our backbone network for ImageNet-LT.
To make a fair comparison with~\cite{wang2017learning}, the pre-trained ResNet-152~\cite{he2016deep} was used as the backbone network for Places-LT.
For MS1M-LT, we used the popular pre-trained ResNet-50~\cite{he2016deep} as the backbone network.
To compare with the recent works, we also adopted the two-headed ResNet-50 backbone following~\cite{wang2020long}.

\begin{table*}[t]
\footnotesize
\centering
\begin{tabular}{l|ccccc|cc|cccc}
\Xhline{1pt}
\textbf{Backbone Net} & \multicolumn{11}{c}{\footnotesize{\textbf{MegaFace Identification Rate}}}  \\
ResNet-50 & $\geqslant5$ & $<5$ \& $\geqslant2$ & $<2$ \& $\geqslant1$ & $=0$ & & \multicolumn{2}{c|}{Gender Sub-Groups} & \multicolumn{4}{c}{Ethnicity Sub-Groups} \\
\textbf{Methods} & \textbf{Many} & \textbf{Few} & \textbf{One-shot} & \textbf{Zero-shot} & \textbf{Full Test} & \textbf{Male} & \textbf{Female} & \textbf{Caucasian} & \textbf{Asian} & \textbf{Indian} & \textbf{African} \\
\hline\hline
Plain Model~\cite{he2016deep} & 80.64 & 71.98 & 84.60 & 77.72 & 73.88 & 78.30 & 78.70 & 85.83 & 75.67 & 76.42 & 79.28 \\ 
Range Loss~\cite{zhang2017range} & 78.60 & 71.36 & 83.14 & 77.40 & 72.17 & 78.12 & 77.45 & 86.11 & 74.86 & 75.94 & 76.37 \\ 
Fair Feature~\cite{zemel2013learning} & - & - & - & - & - & 78.23 & 77.61 & 86.34 & 74.97 & 76.25 & 77.62 \\ 
Debiasing~\cite{ramaswamy2021fair} & - & - & - & - & - & 78.73 & 78.85 & 86.36 & 75.89 & 76.90 & 79.77 \\ 
Ours & \textbf{80.82} & \textbf{72.44} & \textbf{87.60} & \textbf{79.50} & \textbf{74.51} & \textbf{79.04} & \textbf{79.08} & \textbf{86.59} & \textbf{76.22} & \textbf{77.05} & \textbf{80.37} \\
\Xhline{1pt}
\end{tabular}
\caption{
\textbf{Benchmarking results on MegaFace.} Our approach achieved the best performance on natural-world datasets when compared to other state-of-the-art methods. Furthermore, our approach had across-board improvements on `male' and `female' gender sub-groups as well as `Caucasian',`Asian', `Indian' and `African' ethnicity sub-groups.}
\label{tab:benchmark_megaface}
\end{table*}

\begin{table*}[t]
\footnotesize
\centering
\begin{tabular}{l|ccc|cccc|cccc}
\Xhline{1pt}
 & \multicolumn{3}{c|}{CIFAR-10-LT (im. ratio = 100)} & \multicolumn{4}{|c|}{CIFAR-100-LT (im. ratio = 100)} & \multicolumn{4}{c}{iNaturalist-18} \\
\textbf{Methods} & \textbf{Many} & \textbf{Medium} & \textbf{Overall} & \textbf{Many} & \textbf{Medium} & \textbf{Few} & \textbf{Overall} & \textbf{Many} & \textbf{Medium} & \textbf{Few} & \textbf{Overall} \\
\hline\hline
Plain Model~\cite{he2016deep} & - & - & 70.4 & 66.1 & 37.3 & 10.6 & 39.1 & \textbf{72.2} & 63.0 & 57.2 & 61.7 \\ 
\hline
CB Focal~\cite{cui2019class} & - & - & 74.6 & - & - & - & 36.0 & - & - & - & 61.1 \\ 
LDAM~\cite{cao2019learning} & 80.5 & 67.0 & 77.0 & 61.5 & 41.7 & 20.2 & 42.0 & - & - & - & 64.6 \\ 
Decoupling~\cite{kang2019decoupling} & - & - & - & 65.7 & 43.6 & 17.3 & 43.2 & 65.0 & 66.3 & 65.5 & 65.9 \\
BBN~\cite{zhou2020bbn} & - & - & 79.8 & - & - & - & 42.6 & 49.4 & 70.8 & 65.3 & 66.3 \\
RIDE~\cite{wang2020long} & - & - & - & 67.9 & 48.4 & 21.8 & 47.0 & 70.2 & 71.3 & 71.7 & 71.4 \\
\hline
Ours & 79.7 & 62.0 & 76.3 & 61.8 & 41.4 & 17.6 & 41.2 & 62.4 & 66.7 & 65.9 & 65.3 \\
Ours$^{\dagger}$ & \textbf{84.1} & \textbf{68.3} & \textbf{80.8} & \textbf{68.0} & \textbf{48.9} & \textbf{22.6} & \textbf{47.4} & 70.5 & \textbf{72.0} & \textbf{72.2} & \textbf{71.9} \\
\Xhline{1pt}
\end{tabular}
\caption{ 
\textbf{Benchmarking results on CIFAR-10-LT, CIFAR-100-LT and iNaturalist-18.} $^{\dagger}$ denotes using the two-headed ResNet-50 backbone following~\cite{wang2020long}.}
\label{tab:benchmark_inaturalist}
\end{table*}

\vspace{2pt}
\noindent
\textbf{Evaluation Metrics.}
The evaluation is on the performance of each method under both the \emph{closed-set} (test set contains no unknown classes) and \emph{open-set} (test set contains unknown classes) settings.
Under each setting, besides the overall top-$1$ classification accuracy~\cite{gidaris2018dynamic} over all classes, we also calculated the accuracy of three disjoint subsets: \emph{many-shot classes} (classes each with over training 100 samples), \emph{medium-shot classes} (classes each with 20$\sim$100 training samples) and \emph{few-shot classes} (classes under 20 training samples).
For the \emph{open-set} setting, the \emph{F-measure} was also reported for a balanced treatment of precision and recall following~\cite{bendale2016towards}.
For determining open classes, the $softmax$ probability threshold was initially set as $0.1$, while a more detailed analysis is provided in Sec.~\ref{sec:analysis}.

% MegaFace-LT: MegaFace~\cite{kemelmacher2016megaface} is one of the largest face recognition benchmarks. It contains 3,530 images from FaceScrub dataset as a probe set and 1M images as a gallery set. The identification task is to find top-1 nearest image from the 1M gallery for each sample in the probe set. Then the identification rate is the mean of hit rates. 
% Since the identities in training set and testing set are non-overlapped, we adopt an indirect way to partition the testing set into subsets with different shots. We approximate the pseudo occurrences of each test sample by counting the number of the similar (similarity greater than $0.7$) training samples. Apart from many-shot, few-shot, one-shot subsets, we also obtain a zero-shot subset, for which we cannot find similar samples in the training set.

\vspace{2pt}
\noindent
\textbf{Comparison Methods.}
We chose for comparison state-of-the-art methods from different fields dealing with the open long-tailed data, including: (1) \emph{metric learning}: Lifted Loss~\cite{oh2016deep}, (2) \emph{hard negative mining}: Focal Loss~\cite{lin2017focal}, (3) \emph{feature regularization}: Range Loss~\cite{zhang2017range}, (4) \emph{few-shot learning}: FSLwF~\cite{gidaris2018dynamic}, (5) \emph{long-tailed modeling}: MetaModelNet~\cite{wang2017learning}, and (6) \emph{open-set detection}: Open Max~\cite{bendale2016towards}. 
We applied these methods on the same backbone networks as ours for a fair comparison.
We also enabled them with class-aware mini-batch sampling~\cite{shen2016relay} for effective learning.
Since Model Regression~\cite{wang2016learning} and MetaModelNet~\cite{wang2017learning} are the most related to our work, we recorded our results along with the numbers reported in the original papers. We also included the recent advances (\eg CB Focal~\cite{cui2019class}, LDAM~\cite{cao2019learning}, Decoupling~\cite{kang2019decoupling}, BBN~\cite{zhou2020bbn}, and RIDE~\cite{wang2020long}) in long-tailed recognition for a comprehensive evaluation.

\begin{table}[t]
\footnotesize
\centering
\begin{tabular}{l|cc}
\Xhline{1pt}
\textbf{Backbone Net} &  \multicolumn{2}{c}{\textbf{closed-set setting}} \\
ResNet-152 &  &  \\
\textbf{Method} & \textbf{Top-$1$ Accuracy} & \textbf{Top-$5$ Accuracy} \\
\hline\hline
Plain Model~\cite{he2016deep} & 48.0 & 77.8 \\
Cost-Sensitive~\cite{huang2016learning} & 52.4 & 81.9 \\
Model Reg.~\cite{wang2016learning} & 54.7 & 82.4 \\
MetaModelNet~\cite{wang2017learning} & 57.3 & 83.6 \\
\hline
Ours & \textbf{58.7} & \textbf{84.2} \\
\Xhline{1pt}
\end{tabular}
\caption{
\textbf{Benchmarking results on SUN-LT.}}
\label{tab:benchmark_sun}
\end{table}

\subsection{Ablation Study}
\label{sec:ablation}

% We investigate the merit of each module in our framework.
% The performance is reported with \emph{open-set} top-$1$ classification accuracy on ImageNet-LT.

\vspace{2pt}
\noindent
\textbf{Effectiveness of the Dynamic Meta-Embedding.}
From Fig.~\ref{fig:ablation} (b), we observe that the combination of the memory feature and concept selector led to large improvements on all three shots, because the obtained memory feature transferred useful visual concepts among classes.
Another observation is that the confidence calibrator is the most effective on few-shot classes.
And the reachability estimation inside the confidence calibrator helped distinguish tail classes from open classes.

\vspace{2pt}
\noindent
\textbf{Effectiveness of the Modulated Attention.}
We observe from Fig.~\ref{fig:ablation} (a) that, compared to medium-shot classes, the modulated attention contributed more to the discrimination between many-shot and few-shot classes in the experiments. Fig.~\ref{fig:ablation} (c) further validates that the modulated attention is more effective than directly applying spatial attention on feature maps.
It implies that adaptive contexts selection is easier to learn than the conventional feature selection.

\vspace{2pt}
\noindent
\textbf{Effectiveness of the Reachability Calibration.}
To further demonstrate the merit of reachability calibration for open-world setting, we conducted additional experiments following the standard settings in~\cite{hendrycks2016baseline, liang2017enhancing} (CIFAR100 + TinyImageNet(resized)). 
% The results are listed in Table~\ref{tab:openset}, where 
Our approach shows favorable performance over standard open-set methods~\cite{hendrycks2016baseline, liang2017enhancing}, as listed in Table~\ref{tab:openset}.

\subsection{OLTR++ Benchmarking Results}

In this section, we extensively evaluate the performance of various representative methods on our OLTR++ benchmarks.

\begin{figure*}[t]
  \centering
  \includegraphics[width=0.7\textwidth]{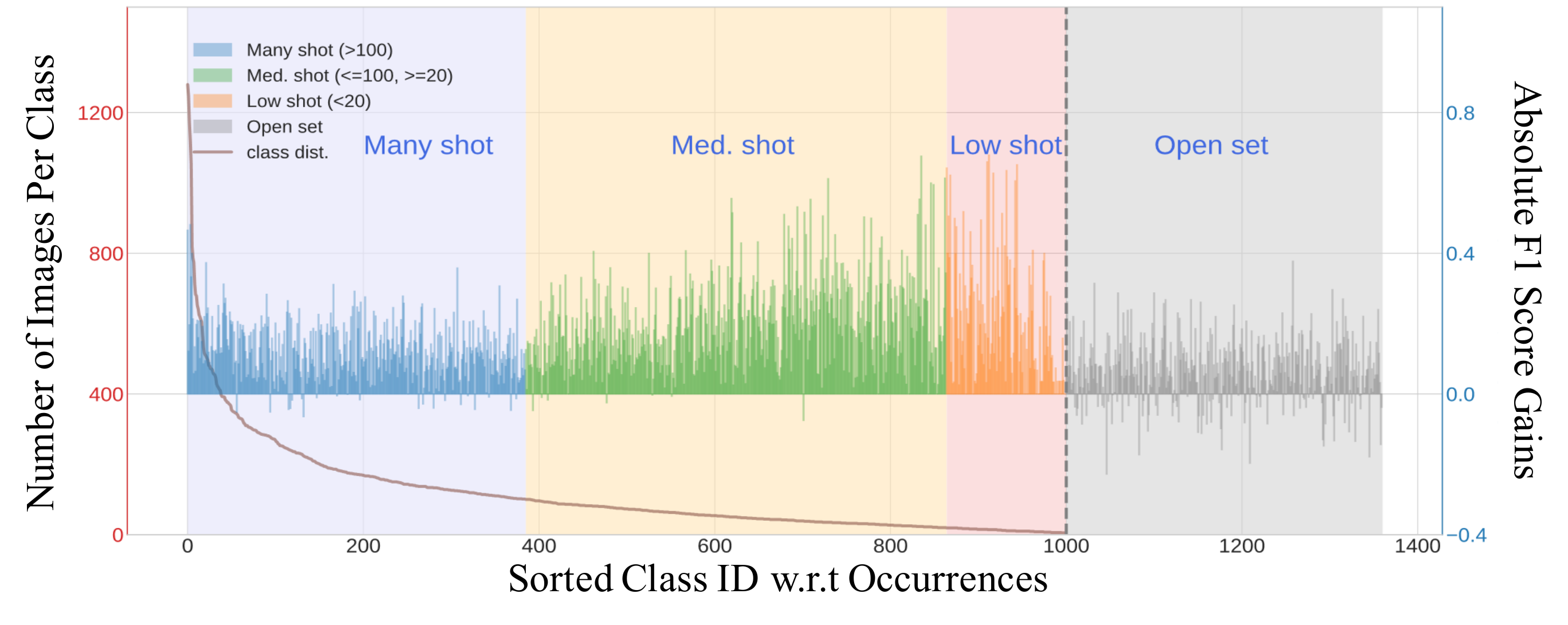}
  \vspace{-8pt}
  \caption{\textbf{The absolute F1 score of our method over the plain model.} Ours has across-the-board performance gains \wrt many/medium/few-shot and open classes.}
  \label{fig:acc_gain}
\end{figure*}

\vspace{2pt}
\noindent
\textbf{ImageNet-LT.}
Table~\ref{tab:benchmark_imagenet} (a) shows the performance comparison of different methods.
We have the following observations.
Firstly, both Lifted Loss~\cite{oh2016deep} and Focal Loss~\cite{lin2017focal} greatly boosted the performance of few-shot classes by enforcing feature regularization.
However, they also sacrificed the performance on many-shot classes since there are no built-in mechanism of adaptively handling  samples of different shots. 
Secondly, OpenMax~\cite{bendale2016towards} improved the results under the open-set setting.
However, the accuracy degraded when it was evaluated with \emph{F-measure}, which considers both precision and recall in open-set.
This proves that when the open classes are compounded with the tail classes, it becomes challenging to perform the distribution fitting that~\cite{bendale2016towards} requires.
Lastly, although the few-shot learning without forgetting approach~\cite{gidaris2018dynamic} retained the many-shot class accuracy, it had difficulty dealing with the imbalanced base classes which are lacked in the current few-shot paradigm.
Overall, as demonstrated in Fig.~\ref{fig:acc_gain}, our approach provides a comprehensive treatment to all the many/medium/few-shot classes as well as the open classes with substantial improvements on all aspects.

\vspace{2pt}
\noindent
\textbf{Places-LT.}
Similar observations can be made on the Places-LT benchmark as shown in Table~\ref{tab:benchmark_imagenet} (b).
With a much stronger baseline (\ie ImageNet pre-trained ResNet-152), our approach still consistently outperformed other alternatives under both the closed-set and open-set settings.
The advantage is even more profound under \emph{F-measure}.

\vspace{2pt}
\noindent
\textbf{MS1M-LT.}
We trained on the MS1M-LT dataset and report results on the MegaFace identification track, which is a standard benchmark in the face recognition field.
Since the face identities in the training set and the test set are disjoint, we adopted an indirect way to partition the testing set into subsets of different shots. We approximated the pseudo shots of each test sample by counting the number of training samples that are similar to it by at least a threshold (feature similarity greater than $0.7$). Apart from many-shot, few-shot, one-shot subsets, we also obtained a zero-shot subset, for which we could not find any sufficiently similar samples in the training set.
It can be observed that our approach has the most advantage on one-shot identities ($3.0\%$ gains) and zero-shot identities ($1.8\%$ gains) as shown in Table~\ref{tab:benchmark_megaface}.

\vspace{2pt}
\noindent
\textbf{SUN-LT.}
To directly compare with~\cite{wang2016learning} and~\cite{wang2017learning}, we also tested out method on the SUN-LT benchmark they provided.
The final results are listed in Table~\ref{tab:benchmark_sun}.
Instead of learning a series of classifier transformations, our approach transfered visual knowledge among features and achieved an $1.4\%$ improvement over the prior best.
Note that our approach also incured much less computational cost since MetaModelNet~\cite{wang2017learning} requires a recursive training procedure.
% for classes with different shots.

\vspace{2pt}
\noindent
\textbf{CIFAR-10-LT~\&~CIFAR-100-LT~\&~iNaturalist-18.}
We also compared OLTR++ with recent proposed methods for long-tailed recognition (\eg CB Focal~\cite{cui2019class}, LDAM~\cite{cao2019learning}, Decoupling~\cite{kang2019decoupling}, BBN~\cite{zhou2020bbn}, and RIDE~\cite{wang2020long}) on CIFAR-10-LT~\cite{cui2019class}, CIFAR-100-LT~\cite{cui2019class} and iNaturalist-18~\cite{van2018inaturalist}.
When combining our framework with the two-headed ResNet-50~\cite{wang2020long}, it achieved state-of-the-art performance on all benchmarks, with consistent gains over many-, medium-, and few-shot classes under various imbalance ratios.

% \begin{table*}[t]
% \footnotesize
% \centering
% \begin{tabular}{l|cccc|cc}
% \Xhline{1pt}
% \textbf{Backbone Net} & \multicolumn{6}{c}{\footnotesize{\textbf{dynamic-set setting}}}  \\
% ResNet-10 & $\geqslant100$ & $<100$ \& $>20$ & $\leqslant20$ & & \multicolumn{2}{c}{Sub-Groups} \\
% \textbf{Methods} & \textbf{Many-shot} & \textbf{Medium-shot} & \textbf{Few-shot} & \textbf{Overall} & \textbf{Known Classes} & \textbf{Unknown Classes} \\
% \hline\hline
% Plain Model~\cite{he2016deep} &  &  &  &  &  &  \\ 
% VAAL~\cite{sinha2019variational} &  &  &  &  &  &  \\ 
% Ours &  &  &  &  &  &  \\
% \Xhline{1pt}
% \end{tabular}
% \caption{ 
% \textcolor{red}{\textbf{Benchmarking results on dynamic-set setting.}}}
% \label{tab:benchmark_active}
% \end{table*}

% \begin{figure*}
%   \centering
%   \includegraphics[width=0.9\textwidth]{figures/active.jpg}
%   \caption{
%   \textcolor{red}{\textbf{Active learning results.}}}
%   \label{fig:active}
% \end{figure*}

\begin{figure*}
  \centering
  \includegraphics[width=1.0\textwidth]{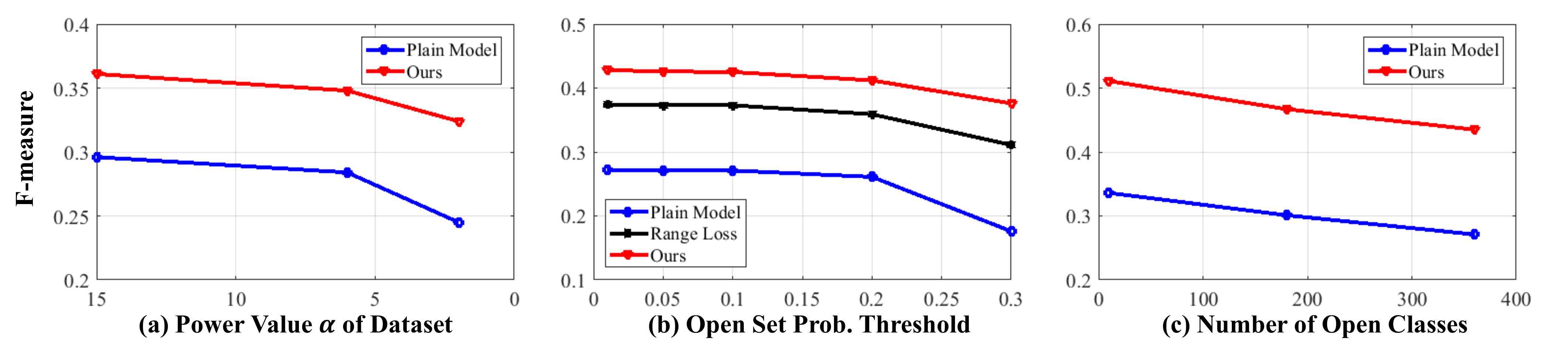}
  \caption{\textbf{The influence of (a) dataset longtail-ness, (b) open-set probability threshold, and (c) the number of open classes.} As the dataset became more imbalanced, our approach only underwent a moderate performance drop. Our approach also demonstrates great robustness to the contamination of open classes.}
  \label{fig:analysis}
\end{figure*}

\subsection{Performance on Fairness Analysis}
\label{sec:fairness}

\vspace{2pt}
\noindent\textbf{Results.}
On the sensitive attribute performance on MS1M-LT, the last six columns in Table~\ref{tab:benchmark_megaface} show that our approach achieved across-board improvements on both gender sub-groups and ethnicity sub-groups, which has an encouraging implication for effective fairness learning.

\vspace{2pt}
\noindent\textbf{Indications.}
The goal of evaluating the performance on different gender sub-groups and ethnicity sub-groups here is to draw connection between long-tailed recognition and the larger community of bias and fairness in artificial intelligence, which could possibly motivate more future research at the intersection of these two.
To provide a comprehensive comparison on this interdisciplinary sub-field, we further implement and include two representative methods in fairness learning (\ie fair feature~\cite{zemel2013learning} and latent debiasing~\cite{ramaswamy2021fair}) into the evaluation.

\begin{table}[t]
\footnotesize
\centering
\begin{tabular}{l|cc}
\Xhline{1pt}
 &  \multicolumn{2}{c}{\textbf{Stage}} \\
 & 1 & 2 \\
 & \textbf{Known~/~Unknown} & \textbf{Known~/~Unknown} \\
\hline\hline
Plain Model~\cite{he2016deep} & 21.4~/~10.8 & 18.2~/~6.6 \\
LwF~\cite{li2017learning} & 26.5~/~17.2 & 24.1~/~11.3 \\
ACL~\cite{ebrahimi2020adversarial} & 30.0~/~21.2 & 27.9~/~16.5 \\
\hline
Ours & \textbf{38.5~/~26.7} & \textbf{36.3~/~24.9} \\
\Xhline{1pt}
\end{tabular}
\caption{
\textbf{Performance of dynamic recognition loop.}}
\label{tab:dynamic_loop}
\end{table}

\subsection{Performance on Active Exploration}
\label{sec:active}

\subsubsection{Effectiveness of Active Sample Selection}

\vspace{2pt}
\noindent\textbf{Setting.}
The experiments were performed on ImageNet-LT (as the initial labeled pool), with the additional classes of images in ImageNet-2010 as the open/exploration set.
This open/exploration set was further mixed with images with known classes, which were taken from the original ImageNet dataset excluding ImageNet-LT.
For fair comparisons, all the methods adopted the same backbone network and were benchmarked under different percentages of selected open samples for oracle annotations (from $10\%$ to $30\%$). 
The evaluation metric was the average recognition accuracy over all the open classes.

\begin{figure}[t]
  \centering
  \includegraphics[width=0.5\textwidth]{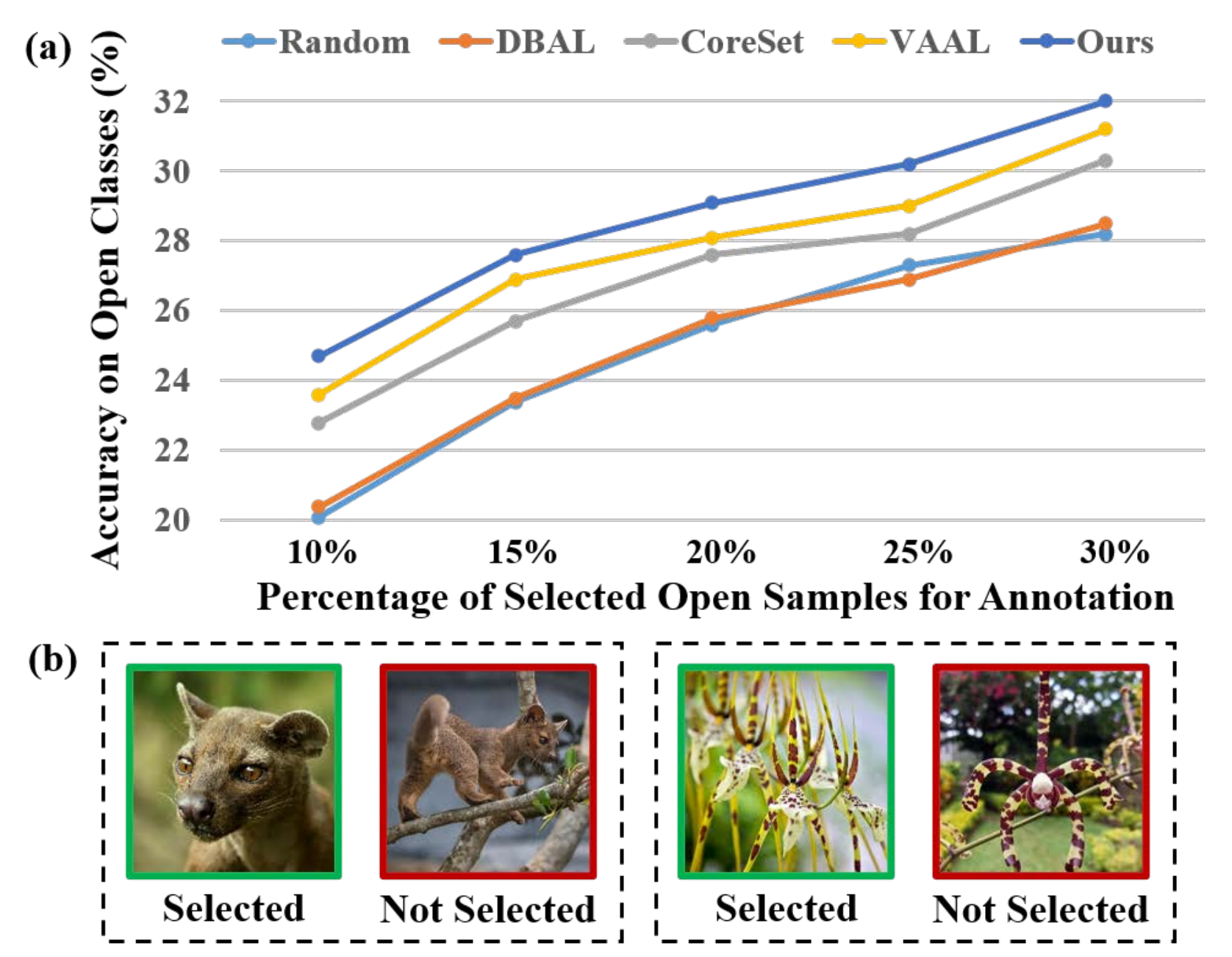}
  \caption{(a) \textbf{Performance of active exploration on open classes.} Our OLTR++ validates its advantage over different percentages of selected open samples for active annotation. (b) \textbf{Samples of selected images} by our active exploration approach.}
  \label{fig:acc_active}
\end{figure}

\vspace{2pt}
\noindent\textbf{Results.}
We evaluated the performance of OLTR++ to recognize the open classes under the active exploration setting. 
We compared our results with several representative methods in active learning, including random sampling, Bayesian-uncertainty-based method DBAL~\cite{gal2017deep}, core-set-based method CoreSet~\cite{sener2017active} and adversarial-learning-based method VAAL~\cite{sinha2019variational}.
As shown in Fig.~\ref{fig:acc_active} (a), our OLTR++ validates its advantage over different percentages of selected open samples for active annotation and demonstrates dynamic-world learning efficiency.

\vspace{2pt}
\noindent\textbf{Analysis.}
To further understand the active exploration ability of OLTR++, we visualized the samples of selected and not selected images by our active exploration approach in Fig.~\ref{fig:acc_active} (b). We can observe that OLTR++ tends to select canonical images with representative parts and appearance for unseen classes.

\subsubsection{Effectiveness of Dynamic Recognition Loop}

\vspace{2pt}
\noindent\textbf{Setting.}
The experiments were performed in a similar setting to active exploration, except that the open/exploration set was divided into two subsets to mimic the two stages in incremental/continual learning scenario.
Specifically, both Stage 1 and Stage 2 have splits from the additional classes of images in ImageNet-2010 as the open/exploration set.
For a fair comparison, the base data is not used for model update in the incremental step.
To mimic the real-world applications, the accuracies of ``known/unknown'' is calculated by classifying test samples to a combination of known and unknown labels.

\vspace{2pt}
\noindent\textbf{Results.}
We compared our results with several representative methods in incremental/continual learning, including Learning without Forgetting (LwF)~\cite{li2017learning}, and Adversarial Continual Learning (ACL)~\cite{ebrahimi2020adversarial}.
LwF~\cite{li2017learning} is a seminal work in the field of incremental learning, which serves as a strong and must-have baseline by employing a soft multi-task learning paradigm. ACL~\cite{ebrahimi2020adversarial} is a recent adversarial-learning-based method of state-of-the-art performance without using external data or models. It is also equipped with open-sourced code and evaluation scripts for fair comparisons.
As listed in Table~\ref{tab:dynamic_loop}, our OLTR++ maintains the learning and recognition effectiveness during the dynamic loop.

\begin{figure*}[t]
  \centering
  \includegraphics[width=1.0\textwidth]{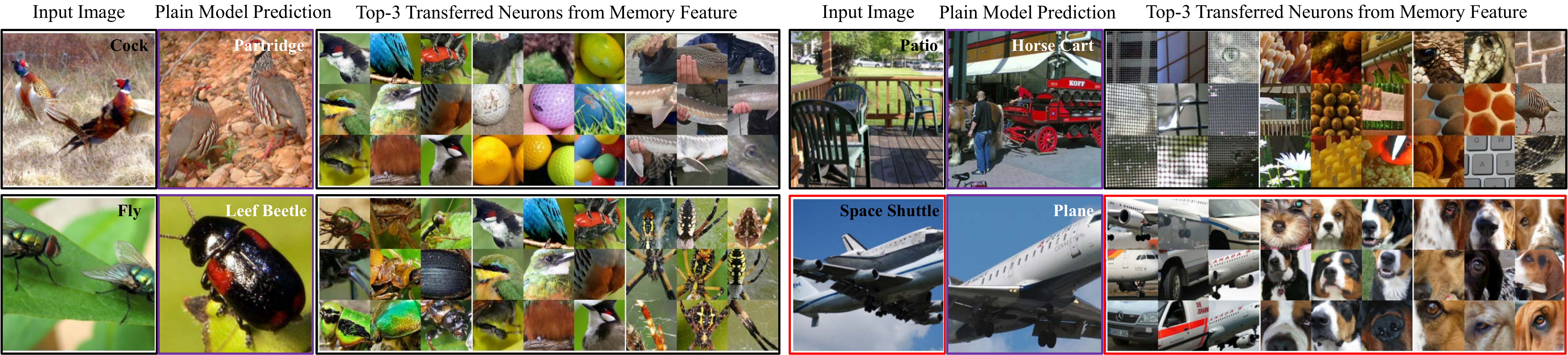}
  \caption{\textbf{Examples of the top-$3$ infused visual concepts from memory feature.} Except for the bottom right failure case 
  % of space shuttle 
  (marked in red), all the other three input images were misclassified by the plain model and correctly classified by our model. For example, to classify the top left image which belongs to a tail class `cock', our approach learned to transfer visual concepts that represents ``bird head'', ``round shape'' and ``dotted texture'' respectively. }
  \label{fig:neuron}
\end{figure*}

\begin{figure*}[t]
  \centering
  \includegraphics[width=1.0\textwidth]{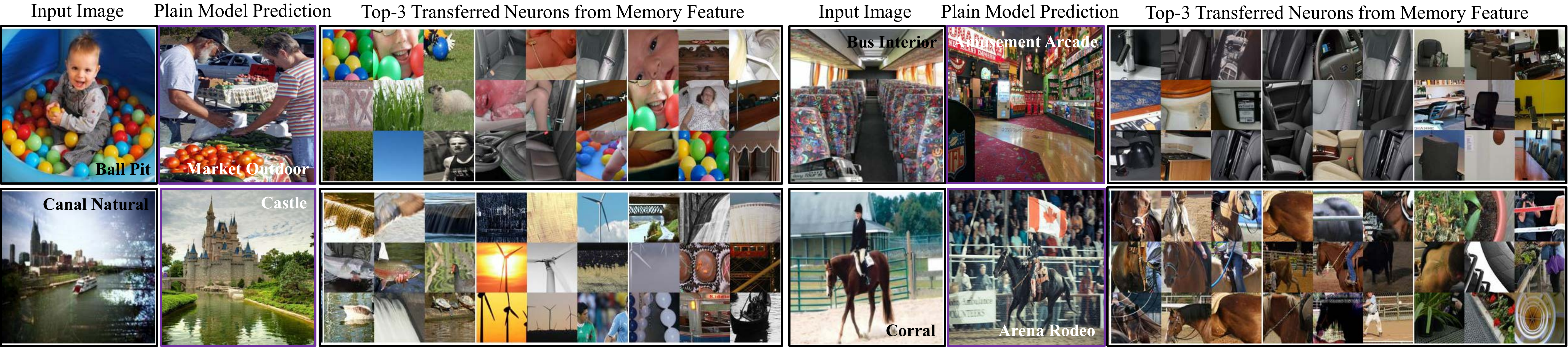}
  \caption{\textbf{Examples of the infused visual concepts from ``memory feature'' in Places-LT.}}
  \label{fig:concept_places}
\end{figure*}

\subsection{Influence of Tailness and Openness}
\label{sec:influence}

% Next we analyze some influencing aspects of tailness and openness to our proposed framework.

\vspace{2pt}
\noindent
\textbf{Influence of Dataset Longtail-ness.}
The longtail-ness of the dataset (\eg the degree of imbalance of the class distribution) can have a negative impact on the model performance. 
For faster investigating, the weights of the backbone network were frozen during training.
From Fig.~\ref{fig:analysis} (a), we observe that as the dataset became more imbalanced (\ie power value $\alpha$ decreases), our approach only underwent a moderate performance drop.
In other words, dynamic meta-embedding enables effective knowledge transfer among data-abundant and data-scarce classes.

\vspace{2pt}
\noindent
\textbf{Influence of Open-Set Probabilistic Threshold.}
The performance change \wrt the open-set probability threshold is demonstrated in Fig.~\ref{fig:analysis} (b).
Compared to the plain model~\cite{he2016deep} and range loss~\cite{zhang2017range}, the performance of our approach changed steadily as the open-set threshold rose.
The reachability estimator in our framework helped calibrate the sample confidence, thus enhanced robustness to open classes.

\vspace{2pt}
\noindent
\textbf{Influence of the Number of Open Classes.}
Finally we investigate performance change \wrt the number of open classes.
Fig.~\ref{fig:analysis} (c) indicates that our approach demonstrates great robustness to the contamination of open classes.

\begin{figure*}[t]
  \centering
  \includegraphics[width=1.0\textwidth]{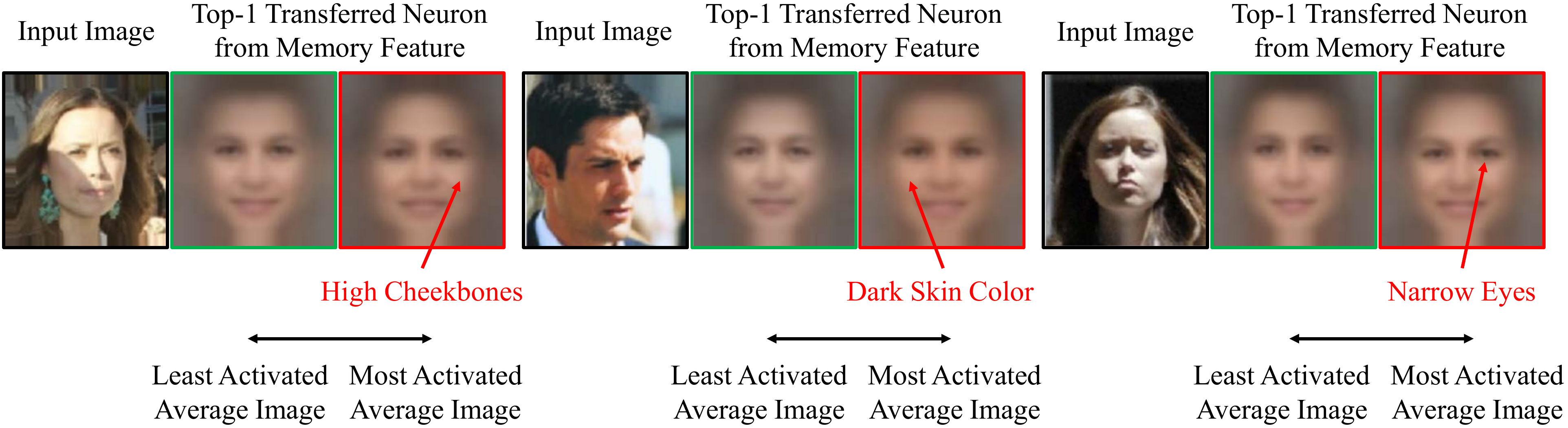}
  \caption{\textbf{Examples of the infused visual concepts from ``memory feature'' in MS1M-LT.}}
  \label{fig:concept_ms1m}
\end{figure*}

\subsection{Further Analysis}
\label{sec:analysis}

In this section, we visualize and interpret memory feature in our framework, as well as present the relation of OLTR++ to fairness analysis and typical failure cases.

\begin{table}[t]
\footnotesize
\centering
\begin{tabular}{l|cc}
\Xhline{1pt}
\textbf{Method} & \textbf{Top-1 Accuracy} & \textbf{Top-5 Accuracy} \\
\hline\hline
Random & 57.9 & 63.2 \\
ImageNet Pre-Train~\cite{deng2009imagenet} & 71.3 & 80.6 \\
Ours & \textbf{77.4} & \textbf{85.8} \\
\Xhline{1pt}
\end{tabular}
\caption{
\textbf{Performance comparison of unsupervised attribute discovery.} It is evaluated on the CelebA dataset~\cite{liu2015deep}.}
\label{tab:attribute}
\end{table}

\vspace{2pt}
\noindent
\textbf{Infused Memory Feature.}
In Fig.~\ref{fig:neuron}, we inspect the visual concepts that memory feature has infused by visualizing its top activating neurons. Specifically, for each input image, we extracted its top-$3$ transferred neurons in memory feature.
And each neuron is visualized by a collection of highest activated patches~\cite{zeiler2014visualizing} over the whole training set. 
For example, to classify the top left image which belongs to a tail class `cock', our approach learned to transfer visual concepts that represents ``bird head'', ``round shape'' and ``dotted texture'' respectively. 
After feature infusion, the dynamic meta-embedding became more informative and discriminative.

We visualize the ``memory feature'' in Places-LT similarly to ImageNet-LT.
Examples of the infused visual concepts from ``memory feature'' in Places-LT are presented in Fig.~\ref{fig:concept_places}.
We observe that ``memory feature'' indeed encodes discriminative visual traits for the underlying scene.

Following~\cite{liu2015deep}, we visualize the ``memory feature'' in MS1M-LT by contrasting the least activated average image and the most activated average image of the top firing neuron.
From Fig.~\ref{fig:concept_ms1m}, we observe that ``memory feature'' in MS1M-LT infused several identity-related attributes (\eg ``high cheekbones'', ``dark skin color'' and ``narrow eyes'') for precise recognition.

\vspace{2pt}
\noindent
\textbf{Unsupervised Attribute Discovery.}
We also quantitatively evaluate the ``memory feature'' in MS1M-LT by performing unsupervised attribute discovery experiment. The performance is reported on the CelebA dataset~\cite{liu2015deep} via linear probing, as listed in Table~\ref{tab:attribute}.
Our approach outperforms randomly initialized and ImageNet pre-trained features on both top-1 and top-5 average accuracy.

% \vspace{2pt}
% \noindent
% \textbf{Fairness Analysis.}
% %
% The open long-tail recognition proposed in our work also has an intrinsic relationship to fairness analysis~\cite{dwork2012fairness, zemel2013learning, madras2018learning, mitchell2018model, anne2018women}. 
% % Their key differences are listed in Table~\ref{tab:differences}.
% On the problem setting side, both open long-tail recognition and fairness analysis aim to tackle the imbalance exists in real-world data.
% Open long-tail recognition focuses on the longtail-ness in both known and unknown categories while fairness analysis deals with the bias in sensitive attributes such as male/female and white/black.

% On the methodology side, both open long-tail recognition and fairness analysis aim to learn transferable representations.
% Open long-tail recognition optimizes for the overall accuracy of all categories while fairness analysis optimizes for several attribute-wise criteria.
% The sensitive attribute (\ie gender and ethnicity sub-groups) results in Table~\ref{tab:benchmark_megaface} demonstrates that our proposed dynamic meta-embedding is also a promising solution to fairness analysis.

\vspace{2pt}
\noindent
\textbf{Failure Cases.}
Since our approach encourages the feature infusion among classes, it slightly sacrifices the fine-grained discrimination for the promotion of under-representative classes. 
One typical failure case of our approach is the confusion between many-shot and medium-shot classes.
For example, the bottom right image in Fig.~\ref{fig:neuron} was misclassified into `airplane' because some cross-category traits like ``nose shape'' and ``eye shape'' were infused.
Feature disentanglement~\cite{bengio2013representation} and strong contrastive learning~\cite{chen2020simple} are potential alleviations to this trade-off issue.

%% file: sections/conclusion.tex
%------------------------------------------------------------------------
\section{Conclusions}

We introduce the OLTR++ task that learns from natural long-tail open-end distributed data and optimizes the overall accuracy over a balanced test set.
% Our key insight is that the traditional deep embedding (``direct feature'') encodes the visual concepts in a slow pace (\ie SGD updates), thus unable to cope with the varying tail/open classes.
We propose an integrated OLTR++ algorithm, dynamic meta-embedding,
% by combining  direct feature and memory feature and by calibrating the reachability of the feature with respect to the visual memory, 
in order to share visual knowledge between head and tail classes and  to reduce confusion between tail and open classes.
% This representation learning is updated in a much faster pace.
We validated our method on three curated large-scale OLTR++ benchmarks (ImageNet-LT, Places-LT and MS1M-LT) as well as active exploration.  
Our work can enable future researches that are directly transferable to real-world applications.

\vspace{2pt}
\noindent
\textbf{Acknowledgements.}
This work is supported by NTU NAP, MOE AcRF Tier 2 (T2EP20221-0033), and under the RIE2020 Industry Alignment Fund – Industry Collaboration Projects (IAF-ICP) Funding Initiative, as well as cash and in-kind contribution from the industry partner(s). This research is also supported by NSF IIS 1835539, Berkeley Deep Drive, DARPA, and US Government fund through Etegent Technologies on Low-Shot Detection in Remote Sensing Imagery. The views, opinions and/or findings expressed are those of the author and should not be interpreted as representing the official views or policies of the Department of Defense or the U.S. Government.

%% file: sections/bio.tex
\begin{IEEEbiography}[{\includegraphics[width=1in,height=1.25in,clip,keepaspectratio]{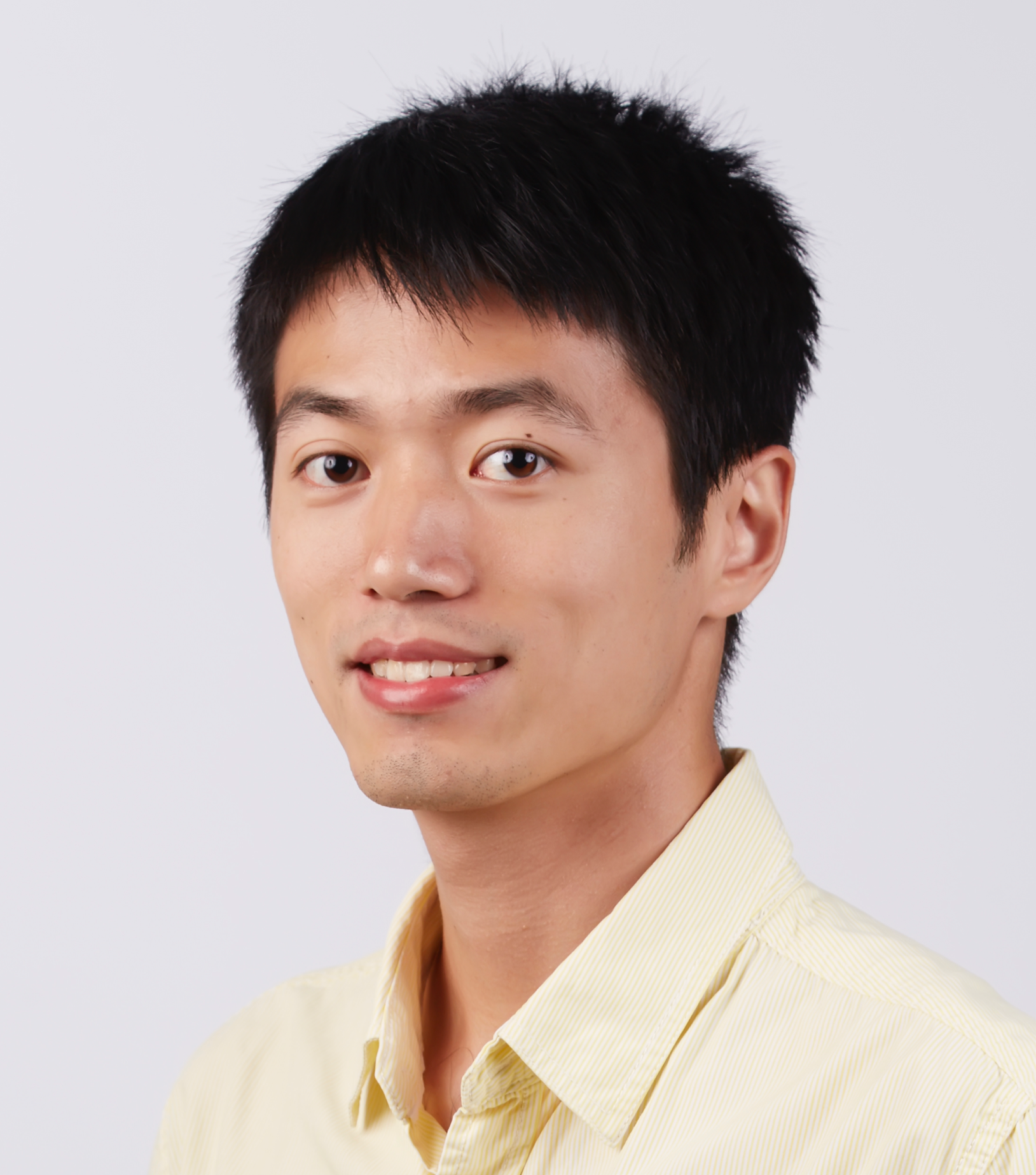}}]{Ziwei Liu}
is currently an Assistant Professor at Nanyang Technological University, Singapore. Previously, he was a senior research fellow at the Chinese University of Hong Kong and a postdoctoral researcher at University of California, Berkeley. Ziwei received his PhD from the Chinese University of Hong Kong. His research revolves around computer vision, machine learning and computer graphics. He has published extensively on top-tier conferences and journals in relevant fields, including CVPR, ICCV, ECCV, NeurIPS, ICLR, ICML, TPAMI, TOG and Nature - Machine Intelligence. He is the recipient of Microsoft Young Fellowship, Hong Kong PhD Fellowship, ICCV Young Researcher Award and HKSTP Best Paper Award. He also serves as an Area Chair of ICCV, NeurIPS and ICLR.
\end{IEEEbiography}

\vskip -2\baselineskip plus -1fil

\begin{IEEEbiography}[{\includegraphics[width=1in,height=1.25in,clip,keepaspectratio]{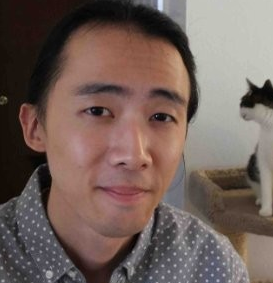}}]{Zhongqi Miao}
 is currently a PhD candidate at International Computer Science Institute, University of California at Berkeley. His research interests include imbalanced classification and domain adaptation in computer vision.
\end{IEEEbiography}

\vskip -2\baselineskip plus -1fil

\begin{IEEEbiography}[{\includegraphics[width=1in,height=1.25in,clip,keepaspectratio]{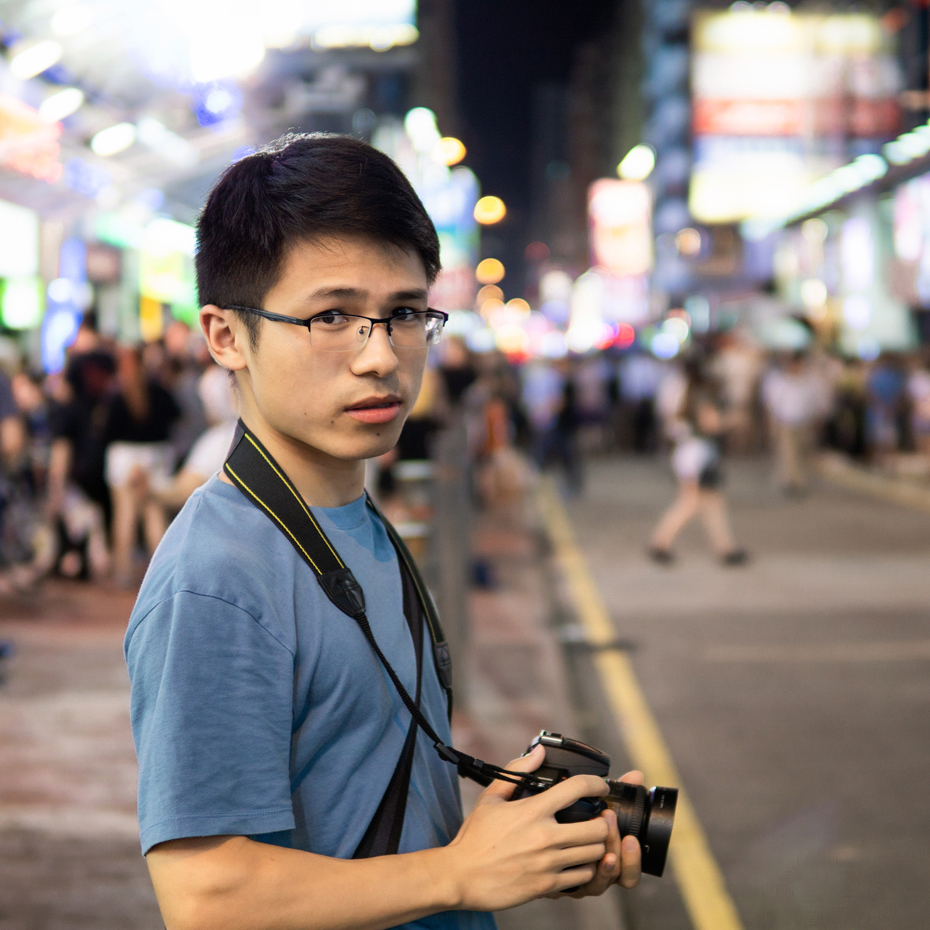}}]{Xiaohang Zhan} is currently a Ph.D. student in Multimedia Laboratory, The Chinese University of Hong Kong. His research interests include computer vision and machine learning, particularly unsupervised learning.
\end{IEEEbiography}

\vskip -2\baselineskip plus -1fil

\begin{IEEEbiography}[{\includegraphics[width=1in,height=1.25in,clip,keepaspectratio]{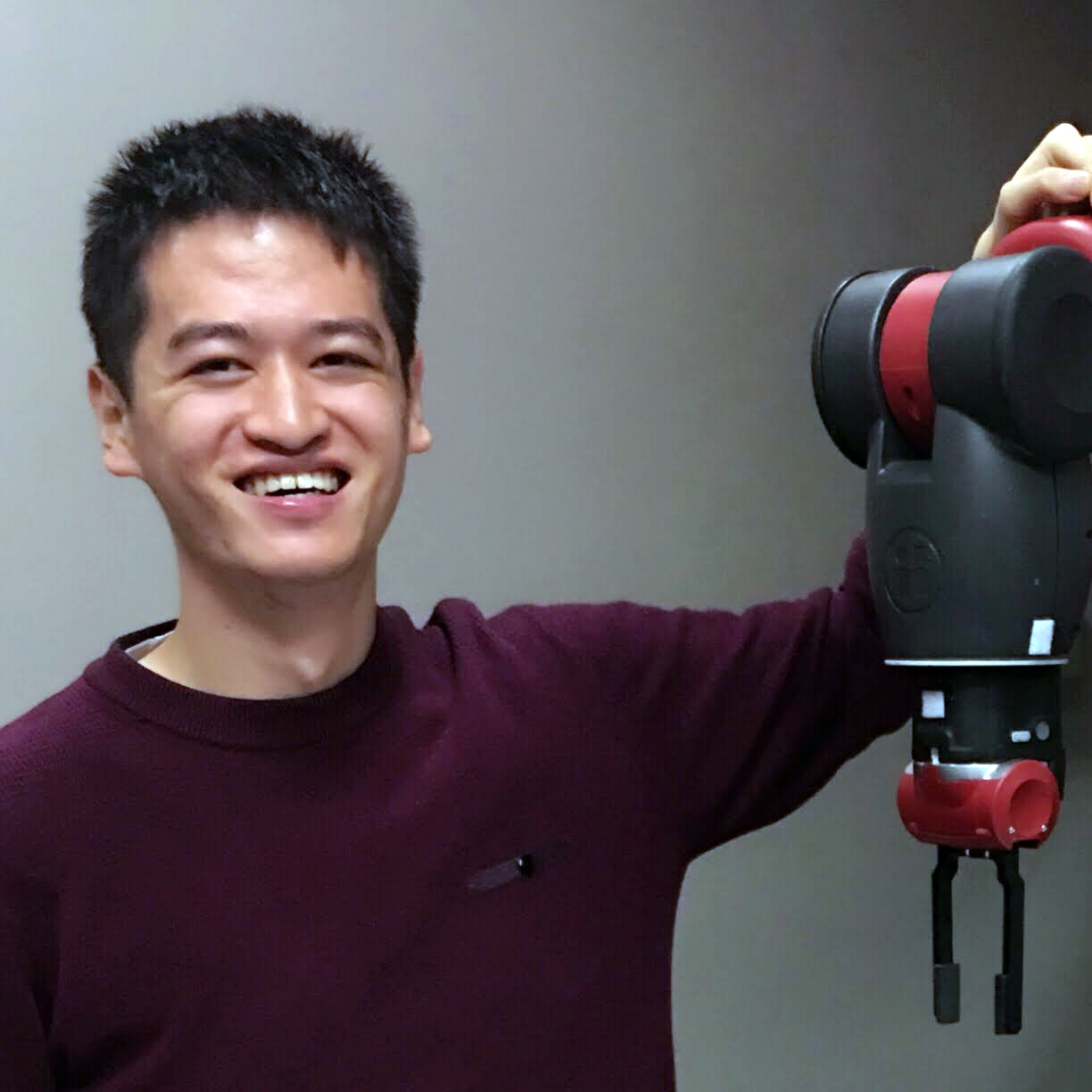}}]{Jiayun Wang} is currently a Ph.D. candidate in the Vision Science program, on computational vision track, University of California at Berkeley. His research interests include 3D vision and unsupervised learning. 
\end{IEEEbiography}

\vskip -2\baselineskip plus -1fil

\begin{IEEEbiography}[{\includegraphics[width=1in,height=1.25in,clip,keepaspectratio]{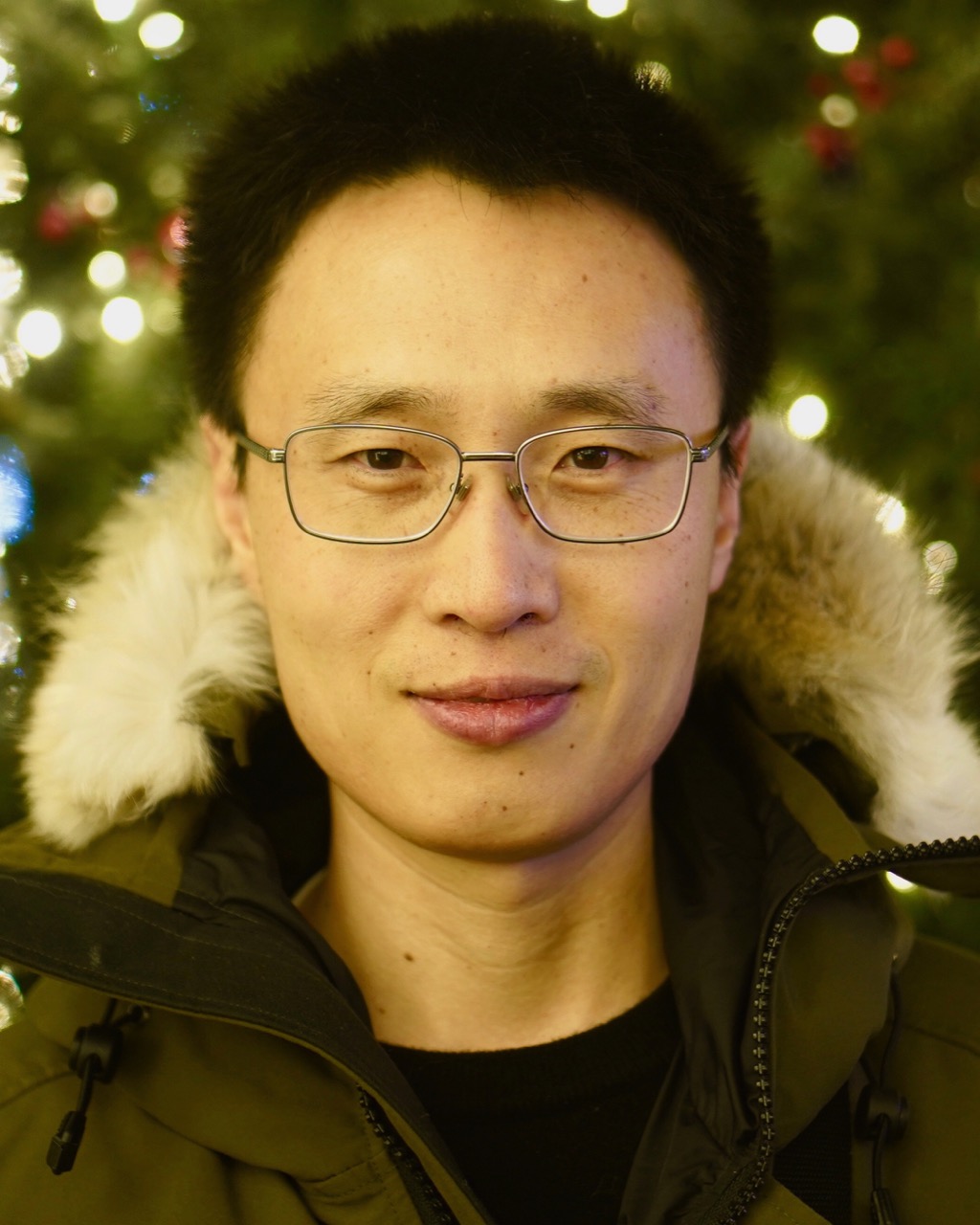}}]{Boqing Gong} received his Ph.D. in 2015 at the University of Southern California, where the Viterbi Fellowship partially supported his work. He is currently a research scientist at Google, Seattle. Before joining Google in 2019, he worked in Tencent AI Lab (Seattle) and was a tenure-track Assistant Professor at the University of Central Florida (UCF). He received an NSF CRII award in 2016 and an NSF BIGDATA award in 2017, both of which were the first of their kinds ever granted to UCF.  His research interestes include data- and label-efficient learning (e.g., domain adaptation, few-shot, reinforcement, webly-supervised, and self-supervised learning), and visual analytics of objects, scenes, human activities, and their attributes.
\end{IEEEbiography}

\vskip -2\baselineskip plus -1fil

\begin{IEEEbiography}[{\includegraphics[width=1in,height=1.25in,clip,keepaspectratio]{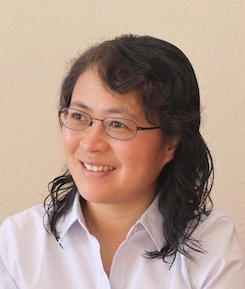}}]{Stella X. Yu} received her Ph.D. from the School of Computer Science at Carnegie Mellon University, where she studied robotics at the Robotics Institute and vision science at the Center for the Neural Basis of Cognition. 
She continued her computer vision research as a postdoctoral fellow at University of California at Berkeley, and then studied art and vision as a Clare Booth Luce Professor at Boston College, during which she received an NSF CAREER award. 
She joined the International Computer Science Institute (ICSI) as a senior research scientist in 2012 and began leading the Vision Group in 2015. She became a Senior Fellow at the Berkeley Institute for Data Science (BIDS) at UC Berkeley in 2016.
Dr. Yu is on the faculty of the Computer Science Department and the Vision Science Program at UC Berkeley, and she is also affiliated faculty with the Department of Computer and Information Science at the University of Pennsylvania. Her research interests include understanding visual perception from multiple perspectives: human vision, computer vision, robotic vision, and artistic vision.
\end{IEEEbiography}

%% file: main.bbl
\begin{thebibliography}{100}
\providecommand{\url}[1]{#1}
\csname url@samestyle\endcsname
\providecommand{\newblock}{\relax}
\providecommand{\bibinfo}[2]{#2}
\providecommand{\BIBentrySTDinterwordspacing}{\spaceskip=0pt\relax}
\providecommand{\BIBentryALTinterwordstretchfactor}{4}
\providecommand{\BIBentryALTinterwordspacing}{\spaceskip=\fontdimen2\font plus
\BIBentryALTinterwordstretchfactor\fontdimen3\font minus
  \fontdimen4\font\relax}
\providecommand{\BIBforeignlanguage}[2]{{%
\expandafter\ifx\csname l@#1\endcsname\relax
\typeout{** WARNING: IEEEtran.bst: No hyphenation pattern has been}%
\typeout{** loaded for the language `#1'. Using the pattern for}%
\typeout{** the default language instead.}%
\else
\language=\csname l@#1\endcsname
\fi
#2}}
\providecommand{\BIBdecl}{\relax}
\BIBdecl

\bibitem{reed2001pareto}
W.~J. Reed, ``The pareto, zipf and other power laws,'' \emph{Economics
  letters}, 2001.

\bibitem{liu2020open}
Z.~Liu, Z.~Miao, X.~Pan, X.~Zhan, D.~Lin, S.~X. Yu, and B.~Gong, ``Open
  compound domain adaptation,'' in \emph{CVPR}, 2020.

\bibitem{deng2009imagenet}
J.~Deng, W.~Dong, R.~Socher, L.-J. Li, K.~Li, and L.~Fei-Fei, ``Imagenet: A
  large-scale hierarchical image database,'' in \emph{CVPR}, 2009.

\bibitem{lin2014microsoft}
T.-Y. Lin, M.~Maire, S.~Belongie, J.~Hays, P.~Perona, D.~Ramanan,
  P.~Doll{\'a}r, and C.~L. Zitnick, ``Microsoft coco: Common objects in
  context,'' in \emph{ECCV}, 2014.

\bibitem{vinyals2016matching}
O.~Vinyals, C.~Blundell, T.~Lillicrap, and D.~Wierstra, ``Matching networks for
  one shot learning,'' in \emph{NeurIPS}, 2016.

\bibitem{lake2015human}
B.~M. Lake, R.~Salakhutdinov, and J.~B. Tenenbaum, ``Human-level concept
  learning through probabilistic program induction,'' \emph{Science}, 2015.

\bibitem{wang2017learning}
Y.-X. Wang, D.~Ramanan, and M.~Hebert, ``Learning to model the tail,'' in
  \emph{NeurIPS}, 2017.

\bibitem{miao2019insights}
Z.~Miao, K.~M. Gaynor, J.~Wang, Z.~Liu, O.~Muellerklein, M.~S. Norouzzadeh,
  A.~McInturff, R.~C. Bowie, R.~Nathan, X.~Y. Stella \emph{et~al.}, ``Insights
  and approaches using deep learning to classify wildlife,'' \emph{Nature -
  Scientific reports}, 2019.

\bibitem{duan2016rl}
Y.~Duan, J.~Schulman, X.~Chen, P.~L. Bartlett, I.~Sutskever, and P.~Abbeel,
  ``Rl$^{2}$: Fast reinforcement learning via slow reinforcement learning,''
  \emph{arXiv preprint arXiv:1611.02779}, 2016.

\bibitem{ba2016using}
J.~Ba, G.~E. Hinton, V.~Mnih, J.~Z. Leibo, and C.~Ionescu, ``Using fast weights
  to attend to the recent past,'' in \emph{NeurIPS}, 2016.

\bibitem{lecun2006tutorial}
Y.~LeCun, S.~Chopra, R.~Hadsell, M.~Ranzato, and F.~Huang, ``A tutorial on
  energy-based learning,'' \emph{Predicting structured data}, vol.~1, no.~0,
  2006.

\bibitem{wang2017non}
X.~Wang, R.~Girshick, A.~Gupta, and K.~He, ``Non-local neural networks,''
  \emph{arXiv preprint arXiv:1711.07971}, 2017.

\bibitem{zhou2018places}
B.~Zhou, A.~Lapedriza, A.~Khosla, A.~Oliva, and A.~Torralba, ``Places: A 10
  million image database for scene recognition,'' \emph{TPAMI}, 2018.

\bibitem{guo2016ms}
Y.~Guo, L.~Zhang, Y.~Hu, X.~He, and J.~Gao, ``Ms-celeb-1m: A dataset and
  benchmark for large-scale face recognition,'' in \emph{ECCV}, 2016.

\bibitem{van2018inaturalist}
G.~Van~Horn, O.~Mac~Aodha, Y.~Song, Y.~Cui, C.~Sun, A.~Shepard, H.~Adam,
  P.~Perona, and S.~Belongie, ``The inaturalist species classification and
  detection dataset,'' in \emph{CVPR}, 2018.

\bibitem{he2016deep}
K.~He, X.~Zhang, S.~Ren, and J.~Sun, ``Deep residual learning for image
  recognition,'' in \emph{CVPR}, 2016.

\bibitem{lin2017focal}
T.-Y. Lin, P.~Goyal, R.~Girshick, K.~He, and P.~Doll{\'a}r, ``Focal loss for
  dense object detection,'' in \emph{ICCV}, 2017.

\bibitem{cui2019class}
Y.~Cui, M.~Jia, T.-Y. Lin, Y.~Song, and S.~Belongie, ``Class-balanced loss
  based on effective number of samples,'' in \emph{CVPR}, 2019.

\bibitem{liu2019large}
Z.~Liu, Z.~Miao, X.~Zhan, J.~Wang, B.~Gong, and S.~X. Yu, ``Large-scale
  long-tailed recognition in an open world,'' in \emph{CVPR}, 2019.

\bibitem{cao2019learning}
K.~Cao, C.~Wei, A.~Gaidon, N.~Arechiga, and T.~Ma, ``Learning imbalanced
  datasets with label-distribution-aware margin loss,'' in \emph{NeurIPS},
  2019.

\bibitem{kang2019decoupling}
B.~Kang, S.~Xie, M.~Rohrbach, Z.~Yan, A.~Gordo, J.~Feng, and Y.~Kalantidis,
  ``Decoupling representation and classifier for long-tailed recognition,'' in
  \emph{ICLR}, 2020.

\bibitem{zhou2020bbn}
B.~Zhou, Q.~Cui, X.-S. Wei, and Z.-M. Chen, ``Bbn: Bilateral-branch network
  with cumulative learning for long-tailed visual recognition,'' in
  \emph{CVPR}, 2020.

\bibitem{wang2020long}
X.~Wang, L.~Lian, Z.~Miao, Z.~Liu, and S.~X. Yu, ``Long-tailed recognition by
  routing diverse distribution-aware experts,'' in \emph{ICLR}, 2021.

\bibitem{salakhutdinov2011learning}
R.~Salakhutdinov, A.~Torralba, and J.~Tenenbaum, ``Learning to share visual
  appearance for multiclass object detection,'' in \emph{CVPR}, 2011.

\bibitem{zhu2014capturing}
X.~Zhu, D.~Anguelov, and D.~Ramanan, ``Capturing long-tail distributions of
  object subcategories,'' in \emph{CVPR}, 2014.

\bibitem{bengio2015the}
S.~Bengio, ``The battle against the long tail,'' in \emph{Talk on Workshop on
  Big Data and Statistical Machine Learning}, 2015.

\bibitem{liu2015deep}
Z.~Liu, P.~Luo, X.~Wang, and X.~Tang, ``Deep learning face attributes in the
  wild,'' in \emph{ICCV}, 2015.

\bibitem{zhu2016we}
X.~Zhu, C.~Vondrick, C.~C. Fowlkes, and D.~Ramanan, ``Do we need more training
  data?'' \emph{IJCV}, 2016.

\bibitem{ouyang2016factors}
W.~Ouyang, X.~Wang, C.~Zhang, and X.~Yang, ``Factors in finetuning deep model
  for object detection with long-tail distribution,'' in \emph{CVPR}, 2016.

\bibitem{liu2016deepfashion}
Z.~Liu, P.~Luo, S.~Qiu, X.~Wang, and X.~Tang, ``Deepfashion: Powering robust
  clothes recognition and retrieval with rich annotations,'' in \emph{CVPR},
  2016.

\bibitem{van2017devil}
G.~Van~Horn and P.~Perona, ``The devil is in the tails: Fine-grained
  classification in the wild,'' \emph{arXiv preprint arXiv:1709.01450}, 2017.

\bibitem{cui2018large}
Y.~Cui, Y.~Song, C.~Sun, A.~Howard, and S.~Belongie, ``Large scale fine-grained
  categorization and domain-specific transfer learning,'' in \emph{CVPR}, 2018.

\bibitem{huang2016learning}
C.~Huang, Y.~Li, C.~C. Loy, and X.~Tang, ``Learning deep representation for
  imbalanced classification,'' in \emph{CVPR}, 2016.

\bibitem{oh2016deep}
H.~Oh~Song, Y.~Xiang, S.~Jegelka, and S.~Savarese, ``Deep metric learning via
  lifted structured feature embedding,'' in \emph{CVPR}, 2016.

\bibitem{dong2017class}
Q.~Dong, S.~Gong, and X.~Zhu, ``Class rectification hard mining for imbalanced
  deep learning,'' in \emph{ICCV}, 2017.

\bibitem{ha2016hypernetworks}
D.~Ha, A.~Dai, and Q.~V. Le, ``Hypernetworks,'' \emph{arXiv preprint
  arXiv:1609.09106}, 2016.

\bibitem{zhang2017range}
X.~Zhang, Z.~Fang, Y.~Wen, Z.~Li, and Y.~Qiao, ``Range loss for deep face
  recognition with long-tailed training data,'' in \emph{CVPR}, 2017.

\bibitem{ye2020identifying}
H.-J. Ye, H.-Y. Chen, D.-C. Zhan, and W.-L. Chao, ``Identifying and
  compensating for feature deviation in imbalanced deep learning,'' \emph{arXiv
  preprint arXiv:2001.01385}, 2020.

\bibitem{menon2020logit}
A.~K. Menon, S.~Jayasumana, A.~S. Rawat, H.~Jain, A.~Veit, and S.~Kumar,
  ``Long-tail learning via logit adjustment,'' \emph{arXiv preprint
  arXiv:2007.07314}, 2020.

\bibitem{wu2020distribution}
T.~Wu, Q.~Huang, Z.~Liu, Y.~Wang, and D.~Lin, ``Distribution-balanced loss for
  multi-label classification in long-tailed datasets,'' in \emph{ECCV}, 2020.

\bibitem{wu2021adversarial}
T.~Wu, Z.~Liu, Q.~Huang, Y.~Wang, and D.~Lin, ``Adversarial robustness under
  long-tailed distribution,'' in \emph{CVPR}, 2021.

\bibitem{ren2022balanced}
J.~Ren, M.~Zhang, C.~Yu, and Z.~Liu, ``Balanced mse for imbalanced visual
  regression,'' in \emph{CVPR}, 2022.

\bibitem{schmidhuber1993neural}
J.~Schmidhuber, ``A neural network that embeds its own meta-levels,'' in
  \emph{ICNN}, 1993.

\bibitem{bertinetto2016learning}
L.~Bertinetto, J.~F. Henriques, J.~Valmadre, P.~Torr, and A.~Vedaldi,
  ``Learning feed-forward one-shot learners,'' in \emph{NeurIPS}, 2016.

\bibitem{ravi2016optimization}
S.~Ravi and H.~Larochelle, ``Optimization as a model for few-shot learning,''
  in \emph{ICLR}, 2017.

\bibitem{santoro2016meta}
A.~Santoro, S.~Bartunov, M.~Botvinick, D.~Wierstra, and T.~Lillicrap,
  ``Meta-learning with memory-augmented neural networks,'' in \emph{ICML},
  2016.

\bibitem{finn2017model}
C.~Finn, P.~Abbeel, and S.~Levine, ``Model-agnostic meta-learning for fast
  adaptation of deep networks,'' \emph{arXiv preprint arXiv:1703.03400}, 2017.

\bibitem{yang2018learning}
F.~S.~Y. Yang, L.~Zhang, T.~Xiang, P.~H. Torr, and T.~M. Hospedales, ``Learning
  to compare: Relation network for few-shot learning,'' in \emph{CVPR}, 2018.

\bibitem{snell2017prototypical}
J.~Snell, K.~Swersky, and R.~Zemel, ``Prototypical networks for few-shot
  learning,'' in \emph{NeurIPS}, 2017.

\bibitem{hariharan2017low}
B.~Hariharan and R.~B. Girshick, ``Low-shot visual recognition by shrinking and
  hallucinating features.'' in \emph{ICCV}, 2017.

\bibitem{wang2018low}
Y.-X. Wang, R.~Girshick, M.~Hebert, and B.~Hariharan, ``Low-shot learning from
  imaginary data,'' \emph{arXiv preprint arXiv:1801.05401}, 2018.

\bibitem{gidaris2018dynamic}
S.~Gidaris and N.~Komodakis, ``Dynamic few-shot visual learning without
  forgetting,'' in \emph{CVPR}, 2018.

\bibitem{ren2018incremental}
M.~Ren, R.~Liao, E.~Fetaya, and R.~S. Zemel, ``Incremental few-shot learning
  with attention attractor networks,'' \emph{arXiv preprint arXiv:1810.07218},
  2018.

\bibitem{qiao2018few}
S.~Qiao, C.~Liu, W.~Shen, and A.~Yuille, ``Few-shot image recognition by
  predicting parameters from activations,'' in \emph{CVPR}, 2018.

\bibitem{qi2018low}
H.~Qi, M.~Brown, and D.~G. Lowe, ``Low-shot learning with imprinted weights,''
  in \emph{CVPR}, 2018.

\bibitem{hinton1987using}
G.~E. Hinton and D.~C. Plaut, ``Using fast weights to deblur old memories,'' in
  \emph{Proceedings of the ninth annual conference of the Cognitive Science
  Society}, 1987.

\bibitem{schmidhuber1992learning}
J.~Schmidhuber, ``Learning to control fast-weight memories: An alternative to
  dynamic recurrent networks,'' \emph{Neural Computation}, 1992.

\bibitem{munkhdalai2017meta}
T.~Munkhdalai and H.~Yu, ``Meta networks,'' \emph{arXiv preprint
  arXiv:1703.00837}, 2017.

\bibitem{scheirer2013toward}
W.~J. Scheirer, A.~de~Rezende~Rocha, A.~Sapkota, and T.~E. Boult, ``Toward open
  set recognition,'' \emph{TPAMI}, 2013.

\bibitem{bendale2016towards}
A.~Bendale and T.~E. Boult, ``Towards open set deep networks,'' in \emph{CVPR},
  2016.

\bibitem{devries2018learning}
T.~DeVries and G.~W. Taylor, ``Learning confidence for out-of-distribution
  detection in neural networks,'' \emph{arXiv preprint arXiv:1802.04865}, 2018.

\bibitem{liang2017enhancing}
S.~Liang, Y.~Li, and R.~Srikant, ``Enhancing the reliability of
  out-of-distribution image detection in neural networks,'' in \emph{ICLR},
  2018.

\bibitem{liu2020energy}
W.~Liu, X.~Wang, J.~D. Owens, and Y.~Li, ``Energy-based out-of-distribution
  detection,'' in \emph{NeurIPS}, 2020.

\bibitem{bendale2015towards}
A.~Bendale and T.~Boult, ``Towards open world recognition,'' in \emph{CVPR},
  2015.

\bibitem{boult2019learning}
T.~E. Boult, S.~Cruz, A.~R. Dhamija, M.~Gunther, J.~Henrydoss, and W.~J.
  Scheirer, ``Learning and the unknown: Surveying steps toward open world
  recognition,'' in \emph{AAAI}, 2019.

\bibitem{joseph2021towards}
K.~Joseph, S.~Khan, F.~S. Khan, and V.~N. Balasubramanian, ``Towards open world
  object detection,'' in \emph{CVPR}, 2021.

\bibitem{lampert2013attribute}
C.~H. Lampert, H.~Nickisch, and S.~Harmeling, ``Attribute-based classification
  for zero-shot visual object categorization,'' \emph{TPAMI}, 2013.

\bibitem{xian2018zero}
Y.~Xian, C.~H. Lampert, B.~Schiele, and Z.~Akata, ``Zero-shot learning—a
  comprehensive evaluation of the good, the bad and the ugly,'' \emph{TPAMI},
  2018.

\bibitem{changpinyo2016synthesized}
S.~Changpinyo, W.-L. Chao, B.~Gong, and F.~Sha, ``Synthesized classifiers for
  zero-shot learning,'' in \emph{CVPR}, 2016.

\bibitem{zhu2017generative}
J.-J. Zhu and J.~Bento, ``Generative adversarial active learning,'' \emph{arXiv
  preprint arXiv:1702.07956}, 2017.

\bibitem{mayer2020adversarial}
C.~Mayer and R.~Timofte, ``Adversarial sampling for active learning,'' in
  \emph{WACV}, 2020.

\bibitem{gal2017deep}
Y.~Gal, R.~Islam, and Z.~Ghahramani, ``Deep bayesian active learning with image
  data,'' \emph{arXiv preprint arXiv:1703.02910}, 2017.

\bibitem{sinha2019variational}
S.~Sinha, S.~Ebrahimi, and T.~Darrell, ``Variational adversarial active
  learning,'' in \emph{ICCV}, 2019.

\bibitem{rusu2016progressive}
A.~A. Rusu, N.~C. Rabinowitz, G.~Desjardins, H.~Soyer, J.~Kirkpatrick,
  K.~Kavukcuoglu, R.~Pascanu, and R.~Hadsell, ``Progressive neural networks,''
  \emph{arXiv preprint arXiv:1606.04671}, 2016.

\bibitem{lee2020neural}
S.~Lee, J.~Ha, D.~Zhang, and G.~Kim, ``A neural dirichlet process mixture model
  for task-free continual learning,'' in \emph{ICLR}, 2019.

\bibitem{rebuffi2017icarl}
S.-A. Rebuffi, A.~Kolesnikov, G.~Sperl, and C.~H. Lampert, ``icarl: Incremental
  classifier and representation learning,'' in \emph{CVPR}, 2017.

\bibitem{aljundi2019gradient}
R.~Aljundi, M.~Lin, B.~Goujaud, and Y.~Bengio, ``Gradient based sample
  selection for online continual learning,'' \emph{NeurIPS}, 2019.

\bibitem{kirkpatrick2017overcoming}
J.~Kirkpatrick, R.~Pascanu, N.~Rabinowitz, J.~Veness, G.~Desjardins, A.~A.
  Rusu, K.~Milan, J.~Quan, T.~Ramalho, A.~Grabska-Barwinska \emph{et~al.},
  ``Overcoming catastrophic forgetting in neural networks,'' \emph{PNAS}, 2017.

\bibitem{zenke2017continual}
F.~Zenke, B.~Poole, and S.~Ganguli, ``Continual learning through synaptic
  intelligence,'' in \emph{ICML}, 2017.

\bibitem{li2017learning}
Z.~Li and D.~Hoiem, ``Learning without forgetting,'' \emph{TPAMI}, 2017.

\bibitem{ebrahimi2020adversarial}
S.~Ebrahimi, F.~Meier, R.~Calandra, T.~Darrell, and M.~Rohrbach, ``Adversarial
  continual learning,'' in \emph{ECCV}, 2020.

\bibitem{dwork2012fairness}
C.~Dwork, M.~Hardt, T.~Pitassi, O.~Reingold, and R.~Zemel, ``Fairness through
  awareness,'' in \emph{The 3rd innovations in theoretical computer science
  conference}, 2012.

\bibitem{zemel2013learning}
R.~Zemel, Y.~Wu, K.~Swersky, T.~Pitassi, and C.~Dwork, ``Learning fair
  representations,'' in \emph{ICML}, 2013.

\bibitem{madras2018learning}
D.~Madras, E.~Creager, T.~Pitassi, and R.~Zemel, ``Learning adversarially fair
  and transferable representations,'' \emph{arXiv preprint arXiv:1802.06309},
  2018.

\bibitem{mitchell2018model}
M.~Mitchell, S.~Wu, A.~Zaldivar, P.~Barnes, L.~Vasserman, B.~Hutchinson,
  E.~Spitzer, I.~D. Raji, and T.~Gebru, ``Model cards for model reporting,''
  \emph{arXiv preprint arXiv:1810.03993}, 2018.

\bibitem{anne2018women}
L.~Anne~Hendricks, K.~Burns, K.~Saenko, T.~Darrell, and A.~Rohrbach, ``Women
  also snowboard: Overcoming bias in captioning models,'' in \emph{ECCV}, 2018.

\bibitem{krizhevsky2012imagenet}
A.~Krizhevsky, I.~Sutskever, and G.~E. Hinton, ``Imagenet classification with
  deep convolutional neural networks,'' in \emph{NeurIPS}, 2012.

\bibitem{hsu2017learning}
Y.-C. Hsu, Z.~Lv, and Z.~Kira, ``Learning to cluster in order to transfer
  across domains and tasks,'' \emph{arXiv preprint arXiv:1711.10125}, 2017.

\bibitem{deng2021joint}
W.~Deng, Q.~Liao, L.~Zhao, D.~Guo, G.~Kuang, D.~Hu, and L.~Liu, ``Joint
  clustering and discriminative feature alignment for unsupervised domain
  adaptation,'' \emph{TIP}, 2021.

\bibitem{wen2016discriminative}
Y.~Wen, K.~Zhang, Z.~Li, and Y.~Qiao, ``A discriminative feature learning
  approach for deep face recognition,'' in \emph{ECCV}, 2016.

\bibitem{savinov2018episodic}
N.~Savinov, A.~Raichuk, R.~Marinier, D.~Vincent, M.~Pollefeys, T.~Lillicrap,
  and S.~Gelly, ``Episodic curiosity through reachability,'' \emph{arXiv
  preprint arXiv:1810.02274}, 2018.

\bibitem{vaswani2017attention}
A.~Vaswani, N.~Shazeer, N.~Parmar, J.~Uszkoreit, L.~Jones, A.~N. Gomez,
  {\L}.~Kaiser, and I.~Polosukhin, ``Attention is all you need,'' in
  \emph{NeurIPS}, 2017.

\bibitem{miao2021iterative}
Z.~Miao, Z.~Liu, K.~M. Gaynor, M.~S. Palmer, S.~X. Yu, and W.~M. Getz,
  ``Iterative human and automated identification of wildlife images,''
  \emph{Nature - Machine Intelligence}, 2021.

\bibitem{zhang2022bamboo}
Y.~Zhang, Q.~Sun, Y.~Zhou, Z.~He, Z.~Yin, K.~Wang, L.~Sheng, Y.~Qiao, J.~Shao,
  and Z.~Liu, ``Bamboo: Building mega-scale vision dataset continually with
  human-machine synergy,'' \emph{arXiv preprint arXiv:2203.07845}, 2022.

\bibitem{sabour2017dynamic}
S.~Sabour, N.~Frosst, and G.~E. Hinton, ``Dynamic routing between capsules,''
  in \emph{NeurIPS}, 2017.

\bibitem{hendrycks2016baseline}
D.~Hendrycks and K.~Gimpel, ``Baseline for detecting misclassified and
  out-of-distribution examples in neural networks,'' in \emph{ICLR}, 2017.

\bibitem{deng2018arcface}
J.~Deng, J.~Guo, and S.~Zafeiriou, ``Arcface: Additive angular margin loss for
  deep face recognition,'' \emph{arXiv preprint arXiv:1801.07698}, 2018.

\bibitem{kemelmacher2016megaface}
I.~Kemelmacher-Shlizerman, S.~M. Seitz, D.~Miller, and E.~Brossard, ``The
  megaface benchmark: 1 million faces for recognition at scale,'' in
  \emph{CVPR}, 2016.

\bibitem{ramaswamy2021fair}
V.~V. Ramaswamy, S.~S. Kim, and O.~Russakovsky, ``Fair attribute classification
  through latent space de-biasing,'' in \emph{CVPR}, 2021.

\bibitem{shen2016relay}
L.~Shen, Z.~Lin, and Q.~Huang, ``Relay backpropagation for effective learning
  of deep convolutional neural networks,'' in \emph{ECCV}, 2016.

\bibitem{wang2016learning}
Y.-X. Wang and M.~Hebert, ``Learning to learn: Model regression networks for
  easy small sample learning,'' in \emph{ECCV}, 2016.

\bibitem{sener2017active}
O.~Sener and S.~Savarese, ``Active learning for convolutional neural networks:
  A core-set approach,'' in \emph{ICLR}, 2018.

\bibitem{zeiler2014visualizing}
M.~D. Zeiler and R.~Fergus, ``Visualizing and understanding convolutional
  networks,'' in \emph{ECCV}, 2014.

\bibitem{bengio2013representation}
Y.~Bengio, A.~Courville, and P.~Vincent, ``Representation learning: A review
  and new perspectives,'' \emph{TPAMI}, 2013.

\bibitem{chen2020simple}
T.~Chen, S.~Kornblith, M.~Norouzi, and G.~Hinton, ``A simple framework for
  contrastive learning of visual representations,'' \emph{arXiv preprint
  arXiv:2002.05709}, 2020.

\end{thebibliography}
